\pdfoutput=1
\documentclass[letterpaper,journal]{IEEEtran}

\usepackage{amsmath,amsfonts}
\usepackage{algorithmic}
\usepackage{algorithm}
\usepackage{array}
\usepackage[caption=false,font=normalsize,labelfont=sf,textfont=sf]{subfig}
\usepackage{textcomp}
\usepackage{stfloats}
\usepackage{placeins}
\usepackage{url}
\usepackage{verbatim}
\usepackage{graphicx}
\usepackage{cite}

\usepackage{xcolor}

\usepackage{amssymb}
\usepackage{mathtools}
\usepackage{amsthm}
\usepackage{multirow}
\usepackage{booktabs}
\usepackage{hyperref}

\DeclareRobustCommand{\rone}[1]{{\color{black}#1}}
\makeatletter
\def\ps@IEEEtitlepagestyle{%
  \let\@oddhead\@empty
  \let\@evenhead\@empty
  \def\@oddfoot{%
    \raisebox{0pt}[0pt][0pt]{%
      \parbox[c]{\textwidth}{%
        \centering\scriptsize\normalfont%
        \textcopyright~2026 IEEE. Personal use of this material is permitted. Permission from IEEE must be obtained for all other uses, in any current or future media, including reprinting/republishing this material for advertising or promotional purposes, creating new collective works, for resale or redistribution to servers or lists, or reuse of any copyrighted component of this work in other works.%
      }%
    }%
  }%
  \def\@evenfoot{\@oddfoot}%
}
\makeatother

\begin{document}
\raggedbottom

\title{UniMap-Verify: One-Class Map-Image Consistency Verification with MLLM-Guided Semantic Hard Negatives}

\author{
Bokang~Yang,
Lichao~Mou,
Yuhang~Yan,
and~Qingyu~Li\textsuperscript{*} ~\IEEEmembership{IEEE Member}

\thanks{ This work was supported by the University Development Fund-Research Start-up Fund from the Chinese University of Hong Kong, Shenzhen (grant NO.UDF01003964).}

\thanks{ B.~Yang, Y.~Yan, and Q.~Li are with School of Science and Engineering, The Chinese University of Hong Kong (Shenzhen), Longgang, Shenzhen, Guangdong, 518172, China (emails: bokangyang@link.cuhk.edu.cn, yuhangyan@link.cuhk.edu.cn, liqingyu@cuhk.edu.cn)}
\thanks{ B.~Yang is also with Shenzhen Loop Area Institute, Shenzhen, 518048, China.}
\thanks{L.~Mou is with Data Science in Earth Observation, Technische Universit{\"a}t M{\"u}nchen (TUM),  Munich, 80333, Germany (email: lichao.mou@tum.de)}
\thanks{\textsuperscript{*}Corresponding author: Q.~Li.}

}

\markboth{UniMap-Verify}%
{Yang \MakeLowercase{\textit{et al.}}: UniMap-Verify}

\maketitle

\begin{abstract}
Ensuring consistency between maps and the real world is critical, yet manual verification remains time-consuming. Aerial imagery offers a cost-effective and widely available proxy for large-scale map verification. In this work, we address unsupervised / one-class map-image consistency verification, where training relies solely on aligned positive pairs without any annotated mismatches. While realistic for map maintenance, this setting is challenging because random negative samples are often too trivial, failing to guide the model toward fine-grained cross-modal verification. To address this problem, we propose UniMap-Verify, a task-focused framework that performs Multimodal Large Language Model (MLLM)-guided semantic hard negative mining. Specifically, we use structured semantic descriptors generated by an MLLM to construct negative samples that are semantically similar to the positives yet geometrically inconsistent with them. A curriculum schedule progressively increases the difficulty of these negatives during training. This strategy shifts the learning focus from coarse semantic shortcuts to geometry-sensitive verification. A  fixed conditional prototype bank is further constructed from matched training pairs to provide complementary normal-reference evidence. Additionally, we introduce MapCheck-Bench, a unified benchmark built from three public sources, comprising approximately  47k  triplets of aerial images, maps, and corresponding MLLM-generated semantic descriptions spanning 42 unique cities or regions. Extensive experiments demonstrate that UniMap-Verify substantially outperforms existing unsupervised baselines under the positive-pair-only protocol. Our code and dataset will be publicly released at \href{https://github.com/yangbokang/UniMap-Verify}{https://github.com/yangbokang/UniMap-Verify}.
\end{abstract}


\section{Introduction}
\IEEEPARstart{W}{ith} rapid urbanization, maps play a crucial role in providing a reference framework that links the physical world to virtual environments. To remain reliable, maps must be continuously updated as changes occur in the real world \cite{fesenmyer2025addressing, ding2020practical}. However, due to the high cost and labor intensity of manual surveying and verification, map updates are typically performed only in response to substantial changes, resulting in notable delays between real-world evolution and its reflection in map data. The increasing availability of high-resolution aerial imagery offers a scalable and cost-effective means of observing the current state of the world, making it possible to support map maintenance with computer vision techniques \cite{zhan2017change}. This gives rise to a fundamental challenge: how to automatically verify whether existing map data remains consistent with the latest aerial imagery.

We formulate this problem as \emph{unsupervised / one-class map-image consistency verification}, which aims to determine whether a given map-image pair is consistent. In this work, we focus on building footprint verification, as urban evolution is primarily manifested through the construction and demolition of buildings \cite{Su2020BUILDINGCD}. This formulation reflects realistic data conditions in map maintenance: matched (positive) pairs corresponding to ``no change'' are abundant, whereas outdated or incorrect (negative) samples are sparse, unpredictable, and costly to annotate at scale. As a result, learning exclusively from positive pairs under a positive-pair-only setting becomes both natural and necessary.

Designing a robust consistency metric in this setting is highly challenging due to the substantial \textit{modality gap} between aerial imagery and maps. Aerial images contain rich textures, illumination variations, and fine-grained visual details, whereas maps represent geographic entities in a sparse, symbolic, and highly abstract manner \cite{deng2024cross}. While human observers can bridge this gap through high-level semantic reasoning—such as recognizing the shape of a stadium despite abstract representation—enabling such reasoning within a learning framework without explicit supervision remains challenging.

More precisely, we adopt a \emph{one-class} training protocol: the model only observes aligned map-image pairs during training, and all mismatched pairs used for optimization are synthesized online as synthetic negatives rather than collected as annotated change examples. We use the term ``unsupervised'' only in this restricted sense, namely that no negative labels or change masks are provided during training. This task definition is therefore stricter than conventional weakly supervised change detection: training data contain only positive pairs, while negatives appear only at evaluation time or as online synthetic mismatches.

A critical examination of existing literature highlights the necessity of a one-class strategy. First, although ``heterogeneous image'' matching is well-studied (e.g., optical vs. SAR) \cite{zhong2022regionclip, sun2021iterative, dong2024contrastive}, the modality gap in those tasks is qualitatively different from the symbolic-to-real gap between maps and aerial imagery, where the latter must bridge abstract geometric primitives with continuous visual signals. Second, map-image matching models \cite{chen2024objformer, bastani2021beyond} typically rely on supervised settings that require large-scale datasets of both matching and non-matching pairs. This requirement creates a bottleneck, as it ignores the real-world scenario where ``changed'' (negative) examples are rare. Therefore, there is an urgent need for a framework capable of learning robust consistency metrics solely from aligned, positive data.

Within this setting, contrastive learning has emerged as a dominant solution for change detection \cite{JSTARS2022HSIChangeDetection, TGRS2023CSSLChangeDetection, CVPR2023ChangeAwareSampling, JSTARS2025PCLCD}. \rone{However, in a one-class verification setting, a central challenge is constructing synthetic negatives from matched pairs to approximate real inconsistencies when real negative samples are unavailable for training.} Conventional contrastive frameworks typically rely on random negative sampling. When applied to map--image verification, these negatives are often \textit{overly easy}, exhibiting obvious semantic mismatches (e.g., pairing an urban map with forest imagery). Models trained in this manner tend to exploit coarse semantic or appearance \textit{shortcuts} rather than learning genuine geometric correspondence. \rone{During training, randomly re-pairing maps and images can produce negatives with obvious differences in building density or scene type. The model can classify these pairs as mismatched using coarse semantic cues alone, reducing the training loss without learning fine-grained spatial correspondence. Such easy negatives therefore provide limited pressure to learn geometric consistency. Selecting negatives from semantically similar scenes reduces the usefulness of these shortcuts and encourages the model to attend to spatial correspondence.} As a result, they generalize poorly to challenging cases where the semantics are similar, but the fine-grained geometry is inconsistent.

To address these limitations, we propose UniMap-Verify, a task-focused framework explicitly designed around semantic hard negatives. Our key insight is that effective one-class verification requires distinguishing pairs that are semantically similar but geometrically inconsistent. We therefore use the zero-shot semantic understanding capability of a Multimodal Large Language Model (MLLM) to extract structured semantic attributes and guide MLLM-driven hard negative mining. On top of that, we introduce a curriculum schedule that progressively increases the proportion of semantic hard negatives, shifting learning from coarse semantic cues to fine-grained geometric discrepancies. A    conditional prototype bank, constructed from matched training pairs, complements this discriminative signal with an explicit normal cross-modal reference.

In addition, we introduce MapCheck-Bench, a unified benchmark that combines three public datasets and enriches them with MLLM-generated semantic descriptors. This benchmark enables rigorous evaluation across diverse urban patterns and geographic regions.

Our contributions are summarized as follows:
\begin{itemize}
\item 	We formalize \emph{unsupervised / one-class map-image consistency verification}, a positive-pair-only task in which training uses only aligned map-image pairs and learns from online synthetic negatives instead of annotated mismatches.
\item 	We propose UniMap-Verify, whose main methodological contribution is MLLM-guided semantic hard negative mining for constructing semantically similar but geometrically inconsistent training pairs.
\item 	We introduce a curriculum schedule that progressively increases hard-negative difficulty and use a conditional prototype bank to store paired map-key/image-value normal references built from matched training pairs only.
\item 	We introduce MapCheck-Bench, a large-scale benchmark built from three public datasets and comprising over    47,000 triplets of maps, aerial images, and corresponding semantic descriptions across 42 unique cities/regions.
\end{itemize}

\begin{figure*}[!t]
  \centering
  \includegraphics[width=1\textwidth]{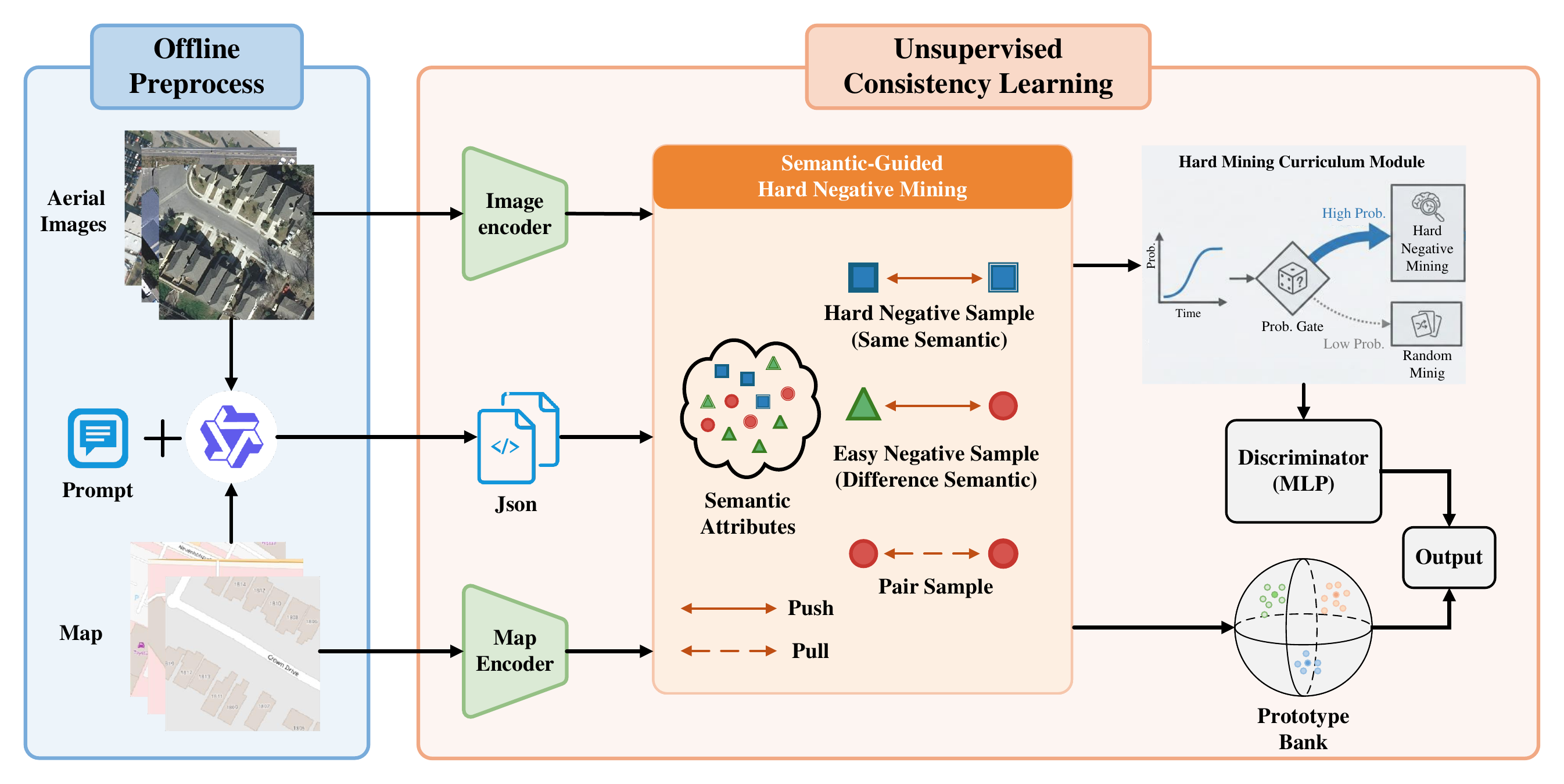}
  \caption{Overview of UniMap-Verify. The framework is centered on one-class map-image consistency verification with MLLM-guided semantic hard negative mining. Negatives are sampled according to semantic attributes derived from MLLM-generated descriptions, and a curriculum schedule progressively increases the proportion of within-group hard negatives. A discriminator provides the primary verification signal, while a conditional map-key/image-value prototype bank supplies complementary evidence learned from matched training pairs.}
  \label{fig:framework}
\end{figure*}

\section{Related Work}
\label{sec:related_work}

Recent research increasingly frames automatic map updating as an incremental refinement problem rather than full reconstruction from scratch. The MUNO21 dataset exemplifies this shift, supporting operations such as adding, removing, and modifying roads while preserving validated regions \cite{bastani2021beyond}. Other works have focused on vector-image alignment and annotation correction, including deep learning methods for cadastral misalignment correction \cite{pisl2021detecting} and the detection of missing buildings in OSM \cite{vargas2019correcting}. General vector-map alignment approaches based on optimization and self-supervision have also been proposed \cite{cherif2024novel}. With the rise of foundation models (e.g., SAM \cite{kirillov2023segment}), zero-shot segmentation has enabled unsupervised discrepancy detection between optical imagery and OpenStreetMap (OSM), scaling map maintenance without extensive supervision \cite{cherif2024novel}. Meanwhile, comprehensive surveys and advanced methods on road extraction and vectorization provide essential methodological context \cite{mo2024survey,sahu2019vector}. Collectively, these works establish a foundation for discriminative frameworks trained only on matching pairs by introducing strategies for alignment, error correction, and handling incomplete annotations.

Unsupervised change detection has also contributed significantly to this area, specifically in tackling the challenges of heterogeneous-sensor analysis, such as optical-SAR data comparison. Many approaches learn modality-invariant representations: for instance, structural relationship graphs with graph-convolutional autoencoders (SR-GCAE) capture structural similarity in multimodal pairs \cite{chen2022unsupervised}, while Markovian co-segmentation jointly infers change masks across heterogeneous modalities \cite{sun2021iterative}. Other methods follow a harmonize-then-compare strategy, using cross-modal translation (with change-aware priors to ignore changed regions \cite{luppino2021deep}) or specialized fusion networks \cite{wang2024cd} to achieve harmonization. Self-supervised contrastive learning is also widely applied: weighted contrastive losses with context-aware Transformers alleviate change leakage in SAR \cite{dong2024contrastive}, radar-optical masked autoencoders learn transferable bimodal features \cite{fuller2023croma}, and semantic-to-change triplets explicitly model temporal differences \cite{ding2025s2c}. While effective for cross-sensor change detection, these approaches are not designed for the map update problem, where the challenge lies in bridging the geometric and semantic gaps between maps and remote sensing imagery.

\section{Methodology}

In this section, we present UniMap-Verify, a task-focused framework for unsupervised / one-class map-image consistency verification. Given a map-image pair, the goal is to determine whether they are mutually consistent, without access to labeled negative samples during training. As illustrated in Figure~\ref{fig:framework}, UniMap-Verify adopts a dual-stream architecture to encode map and image inputs separately,   centers learning on MLLM-guided semantic hard negative mining with curriculum scheduling, and constructs a fixed conditional map-to-image prototype bank from matched training pairs.

Given aligned map-image pairs, where both inputs are rasterized as RGB patches, $x_{map} \in \mathbb{R}^{H \times W \times 3}$ represents the map layer and $x_{img} \in \mathbb{R}^{H \times W \times 3}$ is the aerial imagery, we aim to learn an inconsistency score $A(x_{map}, x_{img}) \in \mathbb{R}$ that is low for corresponding pairs and high for non-corresponding ones. During training, non-corresponding pairs are not annotated data samples; they are synthetic negatives formed by re-pairing matched samples within a mini-batch. This positive-pair-only protocol makes negative construction the central source of supervision.

\rone{Table~\ref{tab:notation} summarizes the notation used in the following method description.}

\begin{table}[t]

\rone{%
\centering
\caption{Main Notations.}
\label{tab:notation}
\small
\setlength{\tabcolsep}{5pt}
\renewcommand{\arraystretch}{1.08}
\begin{tabular}{@{}c l@{}}
\toprule
Notation & Description \\
\midrule
$x_{map},x_{img}$ & Map and aerial-image inputs \\
$\mathcal{D}$ & Matched training set \\
$\mathcal{B},B$ & Mini-batch index set and its size \\
$f_{map},f_{img}$ & Map and image embeddings \\
$f_{joint}$ & Projected joint embedding \\
$m_p,a_p$ & Map and image features at patch $p$ \\
$l_i$ & Semantic label tuple of sample $i$ \\
$\mathcal{H}_i$ & Same-group negative candidate set \\
$p_{hard}(t)$ & Hard-negative probability at epoch $t$ \\
$D$ & Consistency discriminator \\
$\mathcal{P}$ & Conditional prototype bank \\
$K$ & Number of candidate prototype slots \\
$k_j,v_j$ & Map key and image value of slot $j$ \\
$j^*(p)$ & Prototype slot retrieved for patch $p$ \\
$A_{disc,p}$ & Discriminator anomaly score \\
$A_{proto,p}$ & Conditional prototype anomaly score \\
$\widehat{A}_{b,p}$ & Normalized score of branch $b$ \\
$\alpha$ & Discriminator fusion weight \\
$A_{fuse,p}$ & Fused patch anomaly score \\
$A(x)$ & Pair-level inconsistency score \\
$\tau$ & Validation-selected decision threshold \\
\bottomrule
\end{tabular}
\par%
\par}

\end{table}

\subsection{Feature Extraction and Fusion}
We use a shared backbone (pre-trained on ImageNet) to extract multi-scale features from both the map image $x_{map}$ and the aerial image $x_{img}$. Although maps and aerial images are heterogeneous, they share mid-level geometric cues such as footprint boundaries, block layout, and local spatial repetition. Tying the backbone weights encourages both modalities to be projected into a compatible representation space while avoiding unnecessary parameter duplication. Features are extracted from intermediate layers (specifically Layer 2 and Layer 3) to capture both structural detail and scene semantics; relying only on the final stage would discard the fine-grained geometric information needed for footprint verification.

To handle spatial variability, we employ a patch-based processing module. The extracted feature maps are unfolded into patches, interpolated to a fixed reference size, and then processed via an adaptive average pooling layer to obtain compact embeddings    $f_{map} \in \mathbb{R}^{d_f}$ and $f_{img} \in \mathbb{R}^{d_f}$, where $d_f$ denotes the modality-specific feature dimension.

These heterogeneous features are then fused via concatenation and projected into a joint embedding space:
\begin{equation}
    f_{joint} = \text{Projection}(\text{Concat}(f_{map}, f_{img}))
\label{eq:joint_fusion}
\end{equation}
where the Projection module consists of a multi-layer perceptron (MLP) that maps the    $2d_f$-dimensional concatenated input to the target embedding dimension    $d_{embed}$.

\begin{algorithm}[!t]
\caption{Training Procedure of UniMap-Verify}
\label{alg:training}
\begin{algorithmic}

\REQUIRE Aligned map-image pairs with semantic labels $\mathcal{D} = \{(x_{map}^i, x_{img}^i, l_i)\}$, total epochs $T$
\ENSURE    Fixed encoders, a trained projection head and  discriminator, and    a fixed prototype bank
 \STATE Initialize $E_{map}$, $E_{img}$, Projection head,  and  Discriminator $D$
   \STATE \textbf{Step 0: Build Conditional Prototype Bank}
\STATE $\{k_j\}_{j=1}^{K}\leftarrow
       \operatorname{OnePassCluster}(\{\bar m_n\})$
\STATE $c(n)\leftarrow\arg\max_j\;\bar m_n^{\mathsf T}\bar k_j$
\STATE $N_j\leftarrow\lvert\{n:c(n)=j\}\rvert$
\STATE $\mathcal{J}\leftarrow\{j:N_j>0\}$
\STATE $v_j\leftarrow\operatorname{norm}\!\left(
       N_j^{-1}\sum_{c(n)=j}\bar a_n\right),\quad j\in\mathcal{J}$
\STATE $\mathcal{P}\leftarrow\operatorname{Freeze}\!\left(
       \{(k_j,v_j)\}_{j\in\mathcal{J}}\right)$
\STATE \textbf{Step 1: Feature Encoding}
 \FOR{epoch $t = 1$ to $T$}
    \STATE Update curriculum probability $p_{hard}(t)$ via Eq.~\eqref{eq:curriculum}
    \FOR{\rone{each mini-batch $\mathcal{B}$ drawn from $\mathcal{D}$}}
        \STATE   $f_{map}^i \leftarrow E_{map}(x_{map}^i)$, $f_{img}^i \leftarrow E_{img}(x_{img}^i)$
        \STATE $f_{joint}^{real, i} \leftarrow \text{Projection}(\text{Concat}(f_{map}^i, f_{img}^i))$

        \STATE \textbf{Step 2: Semantic-Guided Negative Sampling}

        \STATE \rone{Identify candidate set $\mathcal{H}_i = \{ k \in \mathcal{B} \mid l_k = l_i, k \neq i \}$}
        \STATE Sample negative index $j$:
        \IF{$\mathcal{H}_i \neq \emptyset$ \textbf{and} $\text{Rand}() < p_{hard}(t)$}
            \STATE Sample $j \sim \text{Uniform}(\mathcal{H}_i)$ \quad \textcolor{gray}{\textit{// Hard Negative}}
        \ELSE
            \STATE \rone{Sample $j \sim \text{Uniform}(\{k \in \mathcal{B} \mid k \neq i\})$ \quad \textcolor{gray}{\textit{// Random Negative}}}
        \ENDIF
        \STATE $f_{img}^j \leftarrow E_{img}(x_{img}^j)$
        \STATE $f_{joint}^{fake, i} \leftarrow \text{Projection}(\text{Concat}(f_{map}^i, f_{img}^j))$

        \STATE \textbf{Step 3:    Discriminative  Optimization}
        \STATE Compute $\mathcal{L}_{disc}$    from $f_{joint}^{real}$ and $f_{joint}^{fake}$
         \STATE    Update the projection and discriminator using $\mathcal{L}_{disc}$
     \ENDFOR
\ENDFOR 
\end{algorithmic}
\end{algorithm}

\subsection{Semantic-Guided Hard Negative Mining}
In a one-class setting, synthetic negative construction becomes the main source of supervisory signal. Standard contrastive learning methods typically generate negative pairs via random batch shuffling. This often results in \emph{trivial negatives} where the map and aerial image differ significantly in semantic context (e.g., a high-density urban map paired with a forest image), allowing the model to learn simple statistical shortcuts rather than fine-grained geometric alignment.

To address this, we introduce MLLM-guided Semantic Hard Negative Mining, which is the primary mechanism of UniMap-Verify. We utilize a pre-trained MLLM to generate a structured JSON descriptor for each training sample with four keys: \texttt{density\_class}, \texttt{urban\_style}, \texttt{geographic\_context}, and a free-text \texttt{description}. The first three fields are directly converted into a semantic label tuple $l_i=(d_i,u_i,g_i)$, where $d_i$, $u_i$, and $g_i$ denote \texttt{density\_class}, \texttt{urban\_style}, and \texttt{geographic\_context}, respectively. The free-text description is retained only for qualitative analysis and dataset inspection. Invalid or incomplete outputs are mapped to an \texttt{unknown} group for robustness. This deterministic tuple-based construction avoids an additional embedding-and-clustering stage and makes the grouping rule fully reproducible. \rone{The three attributes constrain complementary coarse properties: \texttt{density\_class} captures building concentration, \texttt{urban\_style} captures broad spatial organization (e.g., grid-like or scattered development), and \texttt{geographic\_context} captures the surrounding environmental setting. Grouping by these image-derived attributes reduces obvious semantic contrasts among candidate scenes.}

During training, given a batch of $B$ positive pairs $\{(x_{map}^i, x_{img}^i)\}_{i=1}^B$, we construct synthetic negative pairs $(x_{map}^i, x_{img}^j)$ ($i \neq j$). We employ a curriculum learning strategy to sample index $j$. Specifically, with probability $p_{hard}(t)$, we sample $j$ from the same semantic group as $i$ (i.e., $l_i = l_j$), creating a \emph{hard negative} that shares high-level semantics but lacks geometric correspondence. With probability    $1-p_{hard}(t)$, we sample $j$ randomly from the batch to maintain global separability. If no same-group candidate exists in the current mini-batch, the sampler falls back to a random negative, as summarized in Algorithm~\ref{alg:training}.

The curriculum probability $p_{hard}$ increases linearly  throughout training:
 \begin{equation}
    p_{hard}(t)=p_{start}
    +(p_{end}-p_{start})\frac{t-1}{T-1},
  \label{eq:curriculum}
\end{equation}
where  $p_{start}=0.2$, $p_{end}=0.9$,  $t$ is the current epoch,  and $T$ is the total number of epochs. The model therefore begins with mostly random negatives and progressively shifts toward semantic hard negatives, reaching the maximum hard-negative probability at the final epoch. \rone{At the beginning of training, easier random mismatches let the projection head and discriminator first learn a basic distinction between matched and mismatched pairs. Increasing the share of same-group negatives then places greater emphasis on spatial correspondence as coarse semantic cues become less informative.}

\subsection{Consistency Learning with a Conditional Prototype Bank }
\rone{UniMap-Verify combines a learned discriminator with a fixed conditional prototype bank. The discriminator learns from matched pairs and synthetic mismatches. The prototype bank stores image references associated with map patterns in matched training pairs and contributes a separate score during inference; it introduces no additional training objective.}

\textbf{Discriminator Branch} The discriminator $D(\cdot)$ is an MLP designed to distinguish between real (positive) pairs and fake (negative) pairs. We employ a Hinge Loss to optimize the discriminator:
\begin{equation}
\begin{aligned}
    \mathcal{L}_{disc} = \frac{1}{B} \sum_{i=1}^B \biggl( &\max(0, m_{disc} - D(f_{joint}^{real, i})) \\
    &+ \max(0, m_{disc} + D(f_{joint}^{fake, i})) \biggr)
\end{aligned}
\end{equation}
where $f_{joint}^{real, i}$ is the fused feature of the matched pair $(x_{map}^i, x_{img}^i)$, and $f_{joint}^{fake, i}$ is the fused feature of the negative pair $(x_{map}^i, x_{img}^j)$. $m_{disc}$ is a margin hyperparameter. This loss encourages the discriminator to output high scores for consistent pairs and low scores for inconsistent ones.

   To align the discriminator output with the anomaly-score convention, its output for each spatial patch $p$ is converted as
\begin{equation}
    A_{disc,p}=1-\sigma\!\left(D(f_{joint,p})\right),
    \label{eq:disc_patch}
\end{equation}
where $D(\cdot)$ denotes the trained discriminator and $\sigma(\cdot)$ is the sigmoid function. Thus, a lower discriminator output produces a higher anomaly-oriented score.

\rone{\textbf{Conditional Prototype Bank.} Our design uses the map to select the image reference against which the observed image is evaluated. Each entry stores a map prototype $k_j\in\mathbb{R}^{d_f}$ as its key and a corresponding image prototype $v_j\in\mathbb{R}^{d_f}$ as its value. Here, a normal reference is built from matched map--image training pairs. Given a query pair, its map feature retrieves a key, and its image feature is compared with the associated value. This provides evidence about whether the image agrees with the expected content for that map pattern, complementing the discriminator trained on synthetic mismatches. The bank uses modality-specific patch features $m_n$ and $a_n$ before pair projection. We allocate $K$ candidate slots and use $K=128$ by default; Section~\ref{subsec:prototype_analysis} evaluates this choice.}

\rone{The bank is constructed once from matched training pairs before discriminator optimization, using features from the frozen encoders.} Let $\bar{u}=u/\lVert u\rVert_2$. \rone{Construction has two stages: the first builds the map keys, and the second computes their associated image values. Map keys are initialized from normalized training map-patch features. In the first stage, each map patch is assigned to its most similar current key.}

\begin{equation}
    \rone{c(n)=\arg\max_{1\leq j\leq K}\;\bar{m}_{n}^{\mathsf T}\bar{k}_{j},}
    \label{eq:pt_key_assignment}
\end{equation}
\rone{and each active key is updated using the normalized cumulative mean of the map features assigned to it,}

\begin{equation}
    \rone{k_j \leftarrow \operatorname{norm}\!\left(
     \frac{1}{N_j}\sum_{n:c(n)=j}\bar{m}_n
    \right).}
    \label{eq:pt_key_update}
\end{equation}
\rone{After this stage, the map keys are fixed. They summarize the map patterns used to select image references.}

\rone{In the second stage, the training map patches are reassigned using the final keys. Each image value is computed from the image features paired with the map patches assigned to that key:}

 \begin{equation}
    \rone{v_j=\operatorname{norm}\!\left(
    \frac{1}{N_j}\sum_{n:c(n)=j}\bar{a}_n
    \right),\qquad j\in\mathcal{J},}
    \label{eq:pt_value}
 \end{equation}
\rone{where $N_j=\lvert\{n:c(n)=j\}\rvert$ counts assignments in the second stage and $\mathcal{J}=\{j:N_j>0\}$ contains the active slots. Only these slots are retained, so each key has an image value and no empty-slot average is computed. The keys and values remain fixed during discriminator training and inference; they receive no gradient and introduce no prototype loss.}

\rone{For a query patch pair $(m_p,a_p)$, the map feature alone selects the most similar map key:}

\begin{equation}
    \rone{j^*(p)=\arg\max_{j\in\mathcal{J}}\;\bar{m}_{p}^{\mathsf T}\bar{k}_{j}.}
    \label{eq:pt_retrieval}
\end{equation}
\rone{The query image does not participate in this selection. It is compared with the retrieved image value to obtain the conditional prototype anomaly score:}

\begin{equation}
    \rone{A_{proto,p}=\frac{1-\bar{a}_{p}^{\mathsf T}\bar{v}_{j^*(p)}}{2},}
    \label{eq:pt_anomaly}
\end{equation}
\rone{The score lies in $[0,1]$ and increases as the observed image feature deviates from the reference retrieved by the map. Thus, the bank supplies a map-conditioned image reference without using the query image to choose that reference.}

\subsection{Inference}
\rone{At inference, the discriminator assesses consistency using the distinction learned from matched pairs and synthetic mismatches, while the prototype branch compares the observed image feature with a normal image reference selected by the map. The latter supplies evidence from matched training pairs that complements the learned discrimination; the two scores are combined at each spatial position before pair-level aggregation.}
For each spatial patch $p$, we use the discriminator anomaly score defined in Eq.~\eqref{eq:disc_patch} together with the conditional prototype score in Eq.~\eqref{eq:pt_anomaly}. \rone{The two branches are independently min--max normalized using fixed validation-set statistics. For branch $b\in\{disc,proto\}$, the normalization is}
\begin{equation}
    \rone{\widehat{A}_{b,p}=\frac{A_{b,p}-m_b^{val}}{M_b^{val}-m_b^{val}+\varepsilon},
    \qquad \varepsilon>0,}
    \label{eq:score_normalization}
\end{equation}
\rone{where $m_b^{val}$ and $M_b^{val}$ are the minimum and maximum patch scores for branch $b$ over the validation set. These statistics are fixed during testing, providing a common scale across input pairs. The stabilizer $\varepsilon$ avoids a zero denominator.} The two normalized maps are then fused at every spatial position,
\begin{equation}
    A_{fuse,p}=\alpha\widehat{A}_{disc,p}
    +(1-\alpha)\widehat{A}_{proto,p},
    \label{eq:score_fusion}
\end{equation}
and global max pooling is applied only after this patch-level fusion:
 \begin{equation}
    A(x)=\max_p A_{fuse,p}.
    \label{eq:disc_max}
 \end{equation}
   \rone{which serves as the final inconsistency score for the map-image pair $x=(x_{map},x_{img})$.} A higher score indicates a greater likelihood of inconsistency.
 \section{The MapCheck-Bench Benchmark}
\label{sec:benchmark}

To robustly evaluate the one-class map verification task, we constructed a unified benchmark, MapCheck-Bench. Rather than collecting raw imagery from scratch, our benchmark contribution lies in unifying heterogeneous public sources into a common verification task with consistent OSM pairing, semantic enrichment, split protocol, and label validation. This section details the data collection, annotation, and statistical distribution.

 \begin{figure*}[!t]
    \centering
    \includegraphics[width=0.9\textwidth]{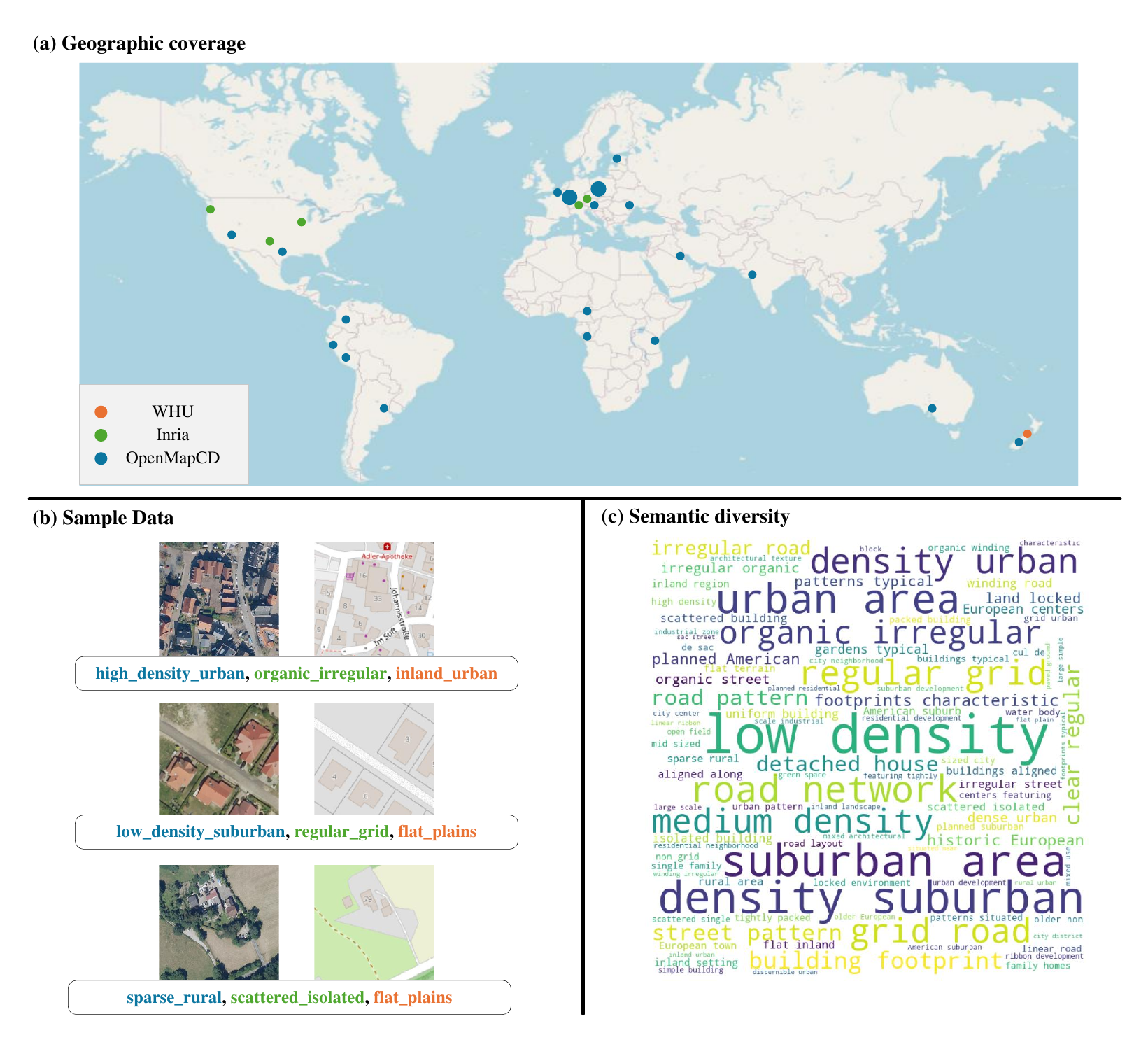}
    \caption{Overview of MapCheck-Bench. The benchmark features (a) geographic coverage across 42 unique cities/regions, (b) sample triplets enriched with MLLM-generated semantic descriptions, and (c) a word cloud illustrating the diverse semantic attributes.}
    \label{fig:Dataset}
\end{figure*}

\subsection{Data Collection}
\label{subsec:data_collection}

Our dataset curation aims for high geographic diversity to ensure the model's robustness across different urban topologies.
After deduplicating overlaps between subsets, the collected data spans 42 unique cities/regions across six continents, including North America, Europe, Asia, Africa, South America, and Oceania.
To intuitively visualize the geographic coverage, we plot the distribution of these cities on a world map, as shown in Figure~\ref{fig:Dataset}(a).

The aerial imagery originates from three primary sources, distinguished by resolution and sensor type:
(1) \textbf{Inria Subset} \cite{maggiori2017dataset} covers 5 regions (e.g., Austin, Vienna) with high-resolution (0.3~m) optical imagery;
(2) \textbf{WHU Subset} \cite{WHUBuilding} covers Christchurch, New Zealand, featuring 0.2~m aerial imagery capturing post-earthquake reconstruction;
(3) \textbf{OpenMapCD Subset} \cite{chen2024objformer} covers 40 diverse regions globally, utilizing OpenEarthMap \cite{xia2023openearthmap} and 0.25-0.5~m ground sampling distance.

Crucially, regardless of the image source, all corresponding vector map data are consistently sourced from OSM \cite{OpenStreetMap}, ensuring the benchmark reflects the real-world challenges of standardizing open-source map data.


\subsection{Data Annotation}
\label{subsec:data_annotation}

The annotation process consists of two stages: geometric ground truth generation and semantic description generation.

\textbf{Change Map Generation.}
To enable quantitative evaluation, we establish pixel-level "Change Maps" that indicate discrepancies between the map and the image (i.e., changed buildings).
For Inria and WHU subsets, we first collect two types of building masks that correspond to aerial imagery and map, respectively. After the collection, two types of building masks are compared in order to label changed buildings between images and maps. 
For OpenMapCD, we utilize existing discrepancy masks between maps and images.
Patch-level changed/unchanged labels are then derived from these masks: a patch is labeled as \emph{changed} if its change mask contains at least one changed-building pixel, and \emph{unchanged} otherwise. All raw data were cropped into non-overlapping patches of \textbf{$256 \times 256$} pixels.

\textbf{Semantic Text Generation.}
Beyond geometric labels, we enrich the dataset with fine-grained semantic descriptions.
We employ Qwen3-VL-8B to generate a structured JSON descriptor for each image patch.
The prompt extracts four fields: \textit{Density Class}, \textit{Urban Style}, \textit{Geographic Context}, and a short free-text \textit{Description}. In the released benchmark, the categorical fields cover six density categories (\textit{sparse\_rural}, \textit{low\_density\_suburban}, \textit{medium\_density}, \textit{high\_density\_urban}, \textit{very\_high\_vertical}, \textit{large\_scale\_industrial}), six urban-style categories (\textit{regular\_grid}, \textit{organic\_irregular}, \textit{linear\_ribbon}, \textit{cul\_de\_sac}, \textit{scattered\_isolated}, \textit{large\_scale\_industrial}), and four geographic-context categories (\textit{inland\_urban}, \textit{coastal\_riparian}, \textit{flat\_plains}, \textit{mountainous\_hilly}). For hard negative mining, we use the tuple of the three categorical fields as the semantic group ID, while the free-text description is reserved for qualitative analysis and word-cloud visualization.
 Before semantic hard-negative sampling, we manually corrected invalid categorical labels based on observable building density, urban layout, and geographic context. Table~\ref{tab:semantic_cleaning} reports 678 corrections among 141,276 fields, corresponding to a pre-correction error rate of 0.480\%.

\begin{table}[!t]
  \centering
  \caption{\textbf{Manual cleaning of MLLM-generated categorical labels.} Values are invalid labels before correction, with field-level error rates in parentheses.}
  \label{tab:semantic_cleaning}
  \setlength{\tabcolsep}{3.5pt}
  \resizebox{\columnwidth}{!}{%
  \begin{tabular}{lrrrr}
    \toprule
    \textbf{Subset}
    & \textbf{Descriptors}
    & \textbf{Density}
    & \textbf{Urban Style}
    & \textbf{Geo. Context} \\
    \midrule
    Inria
    & 24,380
    & 124 (0.509\%)
    & 115 (0.472\%)
    & 119 (0.488\%) \\
    WHU
    & 2,400
    & 10 (0.417\%)
    & 7 (0.292\%)
    & 7 (0.292\%) \\
    OpenMapCD
    & 20,312
    & 98 (0.482\%)
    & 101 (0.497\%)
    & 97 (0.478\%) \\
    \midrule
    \textbf{Total}
    & \textbf{47,092}
    & \textbf{232 (0.493\%)}
    & \textbf{223 (0.474\%)}
    & \textbf{223 (0.474\%)} \\
    \bottomrule
  \end{tabular}
  }
\end{table}

Figure~\ref{fig:Dataset}(b) presents sample data pairs and their corresponding semantic descriptions, while Figure~\ref{fig:mllm_comparison} provides a qualitative comparison of the descriptors generated by different MLLMs.

\subsection{Data Statistics}
\label{subsec:data_statistics}

Based on resolution and data provenance, MapCheck-Bench is organized into three subsets: Inria, WHU, and OpenMapCD.    The  Train/Val/Test ratio is approximately $7{:}1{:}2$  across all subsets. Table~\ref{tab:dataset_stats} summarizes the subset-level statistics.
To rigorously adhere to the one-class, positive-pair-only verification setting:
\begin{itemize}
    \item \textbf{Training Set:} The raw data contain both changed and unchanged pairs. However, for our task we exclusively utilize the unchanged (positive) pairs, ensuring the model strictly learns from aligned data and never observes annotated ``change'' examples during training.
    \item \textbf{Validation \& Test Sets:} These sets contain a balanced ratio ($\approx 1:1$) of changed (negative) and unchanged (positive) pairs. They are used only for model selection and final evaluation, not for constructing the training supervision.
\end{itemize}

 \begin{table}[!t]
    \centering
    \caption{Detailed statistics of MapCheck-Bench    under the approximately $7{:}1{:}2$ train /   validation /   test  split.}
    \label{tab:dataset_stats}
       \resizebox{0.4\textwidth}{!}{
    \begin{tabular}{lcccc}
        \toprule
        \textbf{Subset} & \textbf{Res.} & \textbf{Train} & \textbf{Val} & \textbf{Test} \\
        \midrule
        Inria & 0.3~m & 17,066 & 2,438 & 4,876 \\
        WHU & 0.2~m & 1,680 & 240 & 480 \\
        OpenMapCD & 0.25--0.5~m & 14,249 & 2,031 & 4,032 \\
        \midrule
        \textbf{Total} & -- & \textbf{32,995} & \textbf{4,709} & \textbf{9,388} \\
        \bottomrule
    \end{tabular}
    }
 \end{table}

Additionally, to visualize the semantic diversity of our generated text, we present a Word Cloud in Figure~\ref{fig:Dataset}(c). Terms such as ``dense,'' ``residential,'' ``grid,'' and ``vegetation'' appear frequently, reflecting the dataset's coverage of complex urban environments.

\section{Experiments}
\subsection{Experimental Settings}

 \begin{figure*}[!t]
  \centering
   \includegraphics[width=0.95\textwidth]{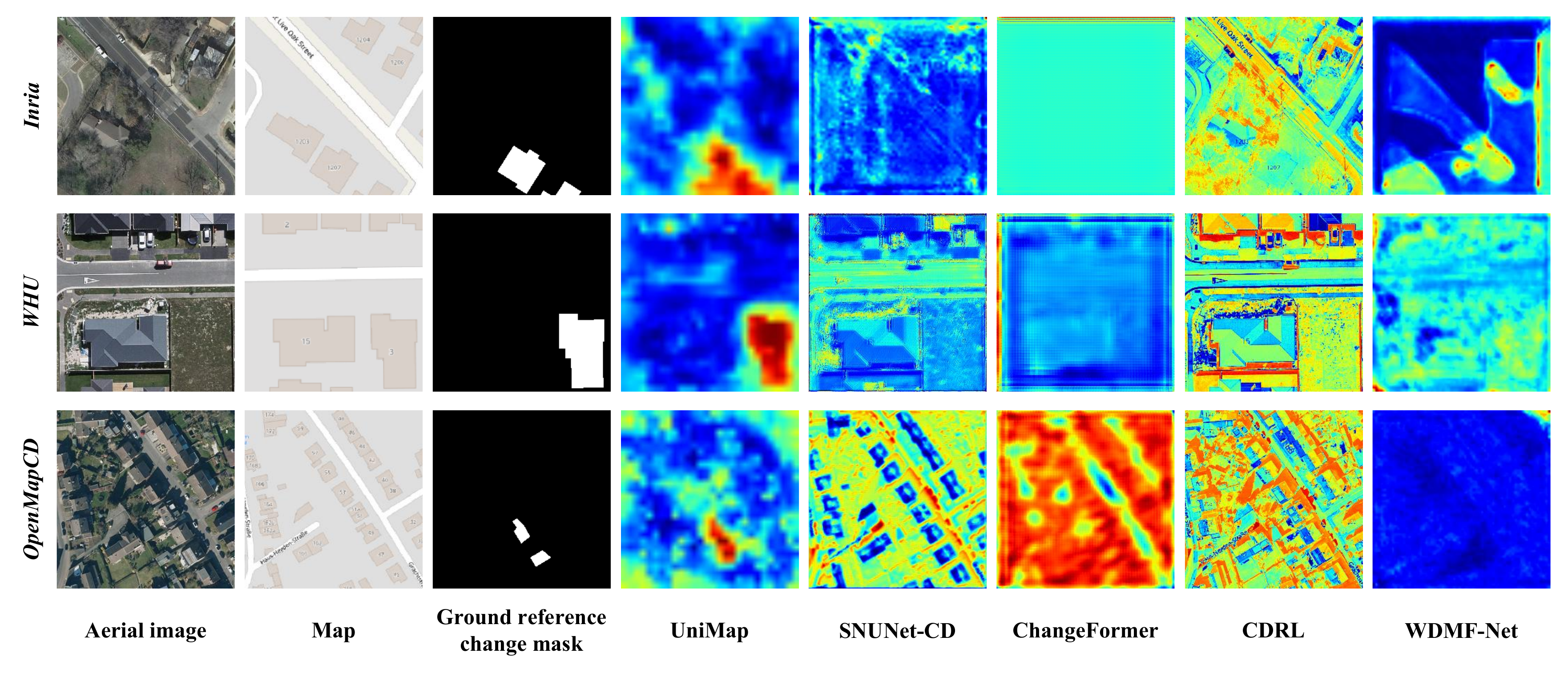}
  \caption{ Qualitative comparison on Inria,  WHU,  and OpenMapCD.  Columns show the aerial image, map, ground-reference change mask, and change-score maps from UniMap-Verify, SNUNet-CD, ChangeFormer, CDRL, and WDMF-Net. Warmer colors indicate higher predicted change scores.}
   \label{fig:results}
\end{figure*}

 \subsubsection{Implementation Details}
All experiments were implemented using PyTorch on a single NVIDIA RTX 5880 Ada GPU. We used WideResNet-50 \cite{zagoruyko2017wideresidualnetworks} pre-trained on ImageNet as a shared, frozen backbone for both image and map encoders, as it provides a stable compromise between receptive field and local-detail preservation for $256 \times 256$ verification patches. The projection head and discriminator were trained for 50 epochs using the AdamW optimizer with an initial learning rate of $1 \times 10^{-4}$ and a weight decay of $1 \times 10^{-2}$. The batch size was set to 128. The hard-negative sampling probability $p_{hard}(t)$ increased linearly from 0.2 to 0.9 according to Eq.~\eqref{eq:curriculum}, and the fusion weight in Eq.~\eqref{eq:score_fusion} was set to $\alpha=0.7$ unless otherwise noted. We use Qwen3-VL-8B \cite{Qwen3-VL} to generate semantic descriptions.

 \begin{figure*}[!t]
  \centering
  \includegraphics[width=0.8\textwidth]{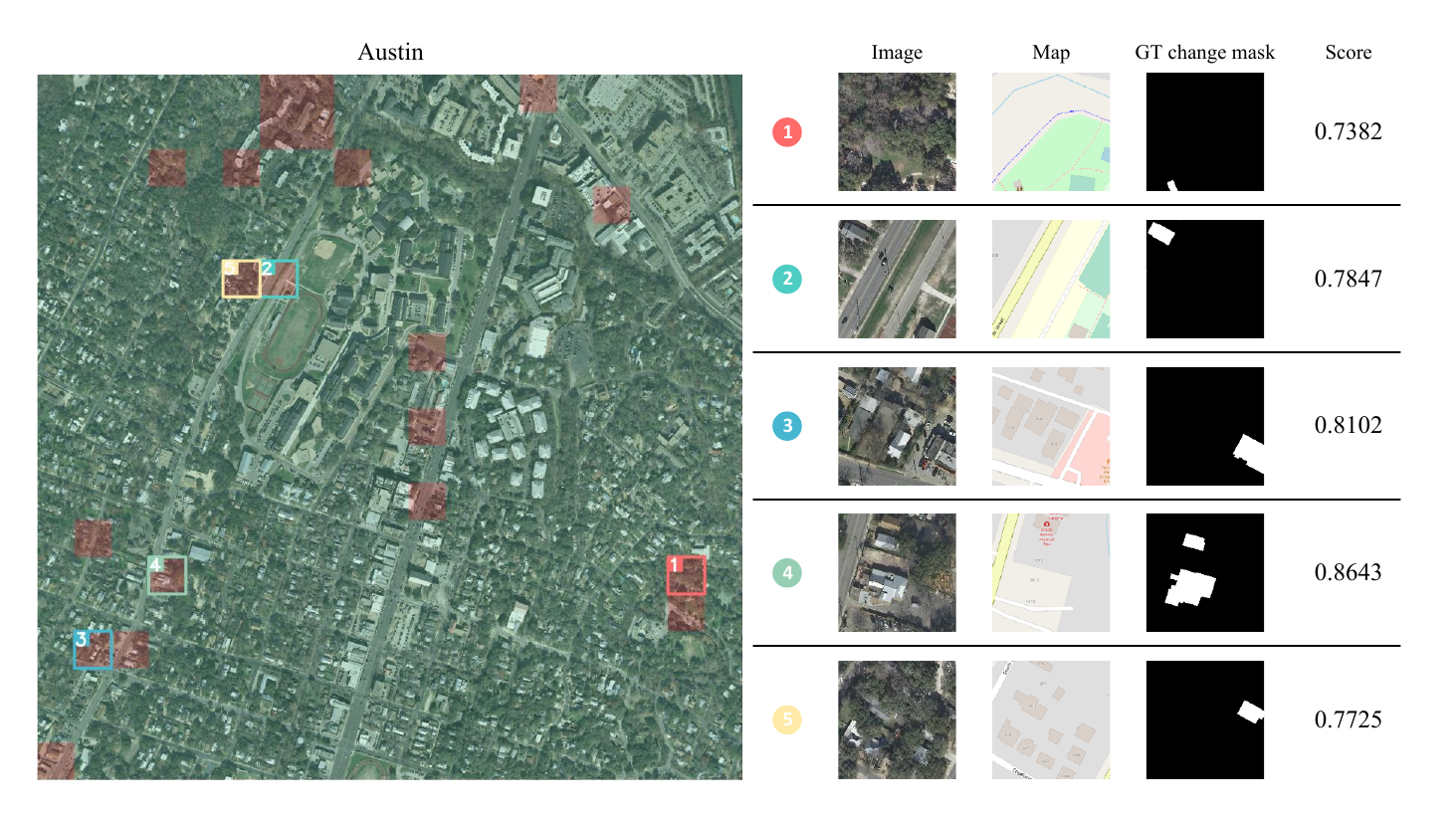}
  \caption{Large-area map verification in Austin (2,250,000 $m^{2}$). Green and red overlays denote predicted consistent and inconsistent regions, respectively. The right panel shows representative locations with their aerial image, map, ground-truth change mask, and image-level inconsistency score.}
  \label{fig:bigcity}
\end{figure*}

 \subsubsection{Evaluation Metrics}
We employ two standard threshold-free metrics to evaluate the performance of verification tasks: Area Under the Receiver Operating Characteristic Curve (AUROC) and Area Under the Precision-Recall Curve (AUPR). Higher values indicate better performance in distinguishing matching pairs from non-matching ones.   To further reflect deployment behavior under a fixed operating point, we also report threshold-based Precision (Pr), Recall (Re), and F1 for the inconsistent/changed class.

\subsubsection{\rone{Experimental Overview}}

\rone{Beyond the main baseline comparisons, geographic and threshold-based evaluations assess generalization to unseen regions and performance (Sections~\ref{subsec:geographic_evaluation} and~\ref{subsec:threshold_evaluation}). Component ablations and prototype analyses isolate the roles of semantic guidance, the reference branch, and the curriculum, and assess sensitivity to prototype capacity and construction (Sections~\ref{subsec:ablation_study} and~\ref{subsec:prototype_bank_analysis}). Negative-sampling comparisons, label-noise tests, feature visualizations, and change-size analysis examine the benefits and robustness of semantic guidance, while descriptor comparisons explain the choice of MLLM (Section~\ref{subsec:semantic_analysis}). Finally, normalization and fusion-weight experiments evaluate how the two inference branches should be combined (Sections~\ref{subsec:score_calibration} and~\ref{subsec:fusion_weight}).}

\subsection{Baselines}
\label{subsec:baselines}
\rone{We evaluate four groups of baselines using matched training pairs without annotated changes. The same group order is used in Tables~\ref{tab:main_results}--\ref{tab:threshold_results}.}

\rone{\textbf{(a) Heterogeneous change detection:} CACD~\cite{wu2021commonality}, CDRL~\cite{CDRL2025}, CFRL~\cite{TIP2025CFRL}, and PRBCD~\cite{hu2023prbcd} detect changes between different remote-sensing modalities without change labels.}

\rone{\textbf{(b) One-class anomaly detection:} Deep SVDD~\cite{ruff2018deep} learns a compact normal representation; PatchCore~\cite{roth2022total} stores normal patch features for anomaly scoring.}

\rone{\textbf{(c) Cross-modal self-supervision:} CMC~\cite{tian2020contrastive} and AVID~\cite{morgado2021audio} provide contrastive cross-modal matching baselines for map--image pairs.}

\rone{\textbf{(d) Adapted supervised change detection:} ObjFormer~\cite{chen2024objformer}, ChangeFormer~\cite{bandara2022transformer}, SEIFNet~\cite{huang2024spatiotemporal}, SNUNet-CD~\cite{fang2021snunet}, WDMF-Net~\cite{liu2026lightweight}, HAM-CD~\cite{li2026ham}, and CDMamba~\cite{zhang2025cdmamba}.} To adapt these supervised baselines to our one-class verification setting, we train each method using only unchanged map-image pairs. For methods that originally output pixel-level change maps, we retain their feature extraction and decoding architectures and supervise them with all-zero change masks on unchanged pairs; at evaluation time, the predicted change-probability map is spatially aggregated into an image-level anomaly score.

\rone{To enable a fairer comparison, we further provide supervised change detection architectures with additional pair-level supervision derived from synthetic mismatches. Specifically, we adapt ChangeFormer, HAM-CD, and CDMamba using aligned map-image pairs as positives and random cross-sample pairings as negatives. The maximum local score is used as the pair-level mismatch score for both training and testing. These variants are denoted by the suffix \textit{-Pair}.}

\begin{table*}[t]

\rone{%
  \centering
  \caption{Comparison on the three datasets (AUROC/AUPR, \%). Groups follow the Baselines subsection; -Pair denotes pair-level adaptation. Best values are in \textbf{bold}.}
  \label{tab:main_results}
  \small\renewcommand{\arraystretch}{0.98}\setlength{\tabcolsep}{5pt}
  \begin{tabular*}{0.8\textwidth}{@{\extracolsep{\fill}}lrrrrrr@{}}
    \toprule
    \multirow{2}{*}{\textbf{Method}} & \multicolumn{2}{c}{\textbf{Inria}} & \multicolumn{2}{c}{\textbf{WHU}} & \multicolumn{2}{c}{\textbf{OpenMapCD}} \\
    \cmidrule(lr){2-3}\cmidrule(lr){4-5}\cmidrule(lr){6-7}
    & AUROC & AUPR & AUROC & AUPR & AUROC & AUPR \\
    \midrule
    \multicolumn{7}{l}{\textit{(a) Heterogeneous change detection}} \\
    CACD & 52.42 & 53.11 & 57.39 & 57.16 & 47.40 & 44.68 \\
    CDRL & 52.69 & 54.85 & 46.04 & 47.80 & 67.27 & 63.03 \\
    CFRL & 48.23 & 50.11 & 41.79 & 43.61 & 49.66 & 46.77 \\
    PRBCD & 50.77 & 52.91 & 43.57 & 45.70 & 65.59 & 61.27 \\
    \midrule
    \multicolumn{7}{l}{\textit{(b) One-class anomaly detection}} \\
    Deep SVDD & 58.31 & 60.17 & 49.81 & 49.58 & 65.27 & 63.62 \\
    PatchCore & 54.52 & 56.01 & 53.78 & 51.88 & 64.23 & 57.93 \\
    \midrule
    \multicolumn{7}{l}{\textit{(c) Cross-modal self-supervision}} \\
    CMC & 58.41 & 61.11 & 60.90 & 63.75 & 59.14 & 62.61 \\
    AVID & 58.44 & 60.92 & 64.84 & 67.54 & 54.45 & 56.09 \\
    \midrule
    \multicolumn{7}{l}{\textit{(d) Adapted supervised change detection}} \\
    ObjFormer & 50.26 & 50.84 & 55.08 & 51.54 & 53.48 & 51.19 \\
    SEIFNet & 51.19 & 52.33 & 57.25 & 58.94 & 38.24 & 41.20 \\
    SNUNet-CD & 48.05 & 49.44 & 54.07 & 53.35 & 50.00 & 49.43 \\
    WDMF-Net & 50.64 & 49.94 & 48.28 & 45.99 & 49.47 & 50.32 \\
    ChangeFormer & 50.16 & 53.20 & 57.98 & 56.87 & 61.04 & 56.86 \\
    ChangeFormer-Pair & 61.76 & 65.40 & 65.32 & 69.27 & 65.23 & 70.08 \\
    HAM-CD & 53.49 & 56.10 & 42.22 & 42.85 & 45.79 & 44.01 \\
    HAM-CD-Pair & 48.04 & 50.68 & 44.54 & 46.55 & 68.60 & 65.70 \\
    CDMamba & 51.90 & 53.01 & 59.92 & 59.32 & 51.25 & 47.77 \\
    CDMamba-Pair & 58.82 & 61.60 & 65.49 & 67.14 & 67.00 & 63.52 \\
    \midrule
    \textbf{UniMap-Verify} & \textbf{64.82} & \textbf{67.88} & \textbf{80.77} & \textbf{82.97} & \textbf{80.20} & \textbf{80.59} \\
    \bottomrule
  \end{tabular*}
\par%
\par}

\end{table*}

\begin{table}[t]

\rone{%
  \centering
  \caption{Geographically disjoint evaluation on OpenMapCD-Split (AUROC/AUPR, \%). Groups follow the Baselines subsection; -Pair denotes pair-level adaptation. Best values are in \textbf{bold}.}
  \label{tab:openmapcd_split_results}
  \fontsize{6}{7.2}\selectfont\renewcommand{\arraystretch}{0.98}\setlength{\tabcolsep}{1pt}
  \begin{tabular*}{\columnwidth}{@{\extracolsep{\fill}}lrrrrrr@{}}
    \toprule
    \multirow{2}{*}{\textbf{Method}} & \multicolumn{2}{c}{\textbf{Asia (incl. Australia)}} & \multicolumn{2}{c}{\textbf{South America}} & \multicolumn{2}{c}{\textbf{Africa}} \\
    \cmidrule(lr){2-3}\cmidrule(lr){4-5}\cmidrule(lr){6-7}
    & AUROC & AUPR & AUROC & AUPR & AUROC & AUPR \\
    \midrule
    \multicolumn{7}{l}{\textit{(a) Heterogeneous change detection}} \\
    CACD & 56.58 & 65.59 & 36.47 & 71.69 & 39.57 & 40.06 \\
    CDRL & 63.96 & 71.28 & 64.06 & 84.70 & 59.66 & 56.11 \\
    CFRL & 42.04 & 57.45 & 62.60 & 87.11 & 39.22 & 38.56 \\
    PRBCD & 62.43 & 69.91 & 62.21 & 83.99 & 54.19 & 47.34 \\
    \midrule
    \multicolumn{7}{l}{\textit{(b) One-class anomaly detection}} \\
    Deep SVDD & 57.98 & 67.30 & 69.33 & 88.94 & 45.49 & 41.49 \\
    PatchCore & 41.65 & 55.48 & 73.00 & 88.40 & 62.66 & 52.86 \\
    \midrule
    \multicolumn{7}{l}{\textit{(c) Cross-modal self-supervision}} \\
    CMC & 66.42 & 72.42 & 36.56 & 72.22 & 27.48 & 33.91 \\
    AVID & 35.83 & 53.49 & 34.78 & 76.82 & 61.25 & 59.91 \\
    \midrule
    \multicolumn{7}{l}{\textit{(d) Adapted supervised change detection}} \\
    ObjFormer & 54.92 & 63.12 & 55.73 & 80.77 & 53.44 & 49.09 \\
    SEIFNet & 51.43 & 61.53 & 48.07 & 78.69 & 36.91 & 36.88 \\
    SNUNet-CD & 60.75 & 70.37 & 66.54 & 89.56 & 50.93 & 47.32 \\
    WDMF-Net & 62.57 & 69.36 & 84.01 & 93.92 & 62.88 & 52.77 \\
    ChangeFormer & 60.25 & 64.94 & 42.49 & 76.78 & 58.48 & 49.89 \\
    ChangeFormer-Pair & 63.30 & 67.09 & 33.61 & 74.33 & 49.67 & 42.73 \\
    HAM-CD & 60.06 & 64.85 & 86.58 & 95.02 & 60.44 & 51.75 \\
    HAM-CD-Pair & 70.95 & 78.98 & 63.93 & 85.70 & 54.87 & 47.91 \\
    CDMamba & 52.31 & 62.85 & 45.15 & 77.24 & 39.14 & 38.42 \\
    CDMamba-Pair & 68.23 & 72.99 & 51.19 & 79.34 & 43.19 & 41.16 \\
    \midrule
    \textbf{UniMap-Verify} & \textbf{79.08} & \textbf{86.21} & \textbf{89.93} & \textbf{97.37} & \textbf{76.70} & \textbf{65.06} \\
    \bottomrule
  \end{tabular*}
\par%
\par}

\end{table}

\begin{table}[t]

\rone{%
  \centering
  \caption{Threshold-based test performance (Precision/Recall/F1, \%). Groups follow the Baselines subsection; -Pair denotes pair-level adaptation. The highest F1 per dataset is \textbf{bold}.}
  \label{tab:threshold_results}
  \fontsize{6}{7.2}\selectfont\renewcommand{\arraystretch}{0.98}\setlength{\tabcolsep}{1pt}
  \begin{tabular*}{\columnwidth}{@{\extracolsep{\fill}}lccc@{}}
    \toprule
    \multirow{2}{*}{\textbf{Method}} & \textbf{Inria} & \textbf{WHU} & \textbf{OpenMapCD} \\
    \cmidrule(lr){2-2}\cmidrule(lr){3-3}\cmidrule(lr){4-4}
    & Pr / Re / F1 & Pr / Re / F1 & Pr / Re / F1 \\
    \midrule
    \multicolumn{4}{l}{\textit{(a) Heterogeneous change detection}} \\
    CACD & 51.37 / 48.70 / 50.00 & 48.10 / 48.10 / 48.10 & 49.62 / 51.00 / 50.30 \\
    CDRL & 59.66 / 53.30 / 56.30 & 43.76 / 49.60 / 46.50 & 57.22 / 62.40 / 59.70 \\
    CFRL & 51.37 / 48.70 / 50.00 & 48.51 / 47.50 / 48.00 & 50.32 / 51.70 / 51.00 \\
    PRBCD & 58.91 / 55.40 / 57.10 & 47.56 / 51.60 / 49.50 & 64.41 / 63.40 / 63.90 \\
    \midrule
    \multicolumn{4}{l}{\textit{(b) One-class anomaly detection}} \\
    Deep SVDD & 57.34 / 57.48 / 57.41 & 48.56 / 64.68 / 55.47 & 57.36 / 72.15 / 63.91 \\
    PatchCore & 55.48 / 43.96 / 49.05 & 50.75 / 72.34 / 59.65 & 57.28 / 88.21 / 69.46 \\
    \midrule
    \multicolumn{4}{l}{\textit{(c) Cross-modal self-supervision}} \\
    CMC & 55.72 / 64.68 / 59.87 & 57.32 / 40.00 / 47.12 & 66.38 / 34.67 / 45.55 \\
    AVID & 59.66 / 40.88 / 48.52 & 61.18 / 44.26 / 51.36 & 55.27 / 33.17 / 41.46 \\
    \midrule
    \multicolumn{4}{l}{\textit{(d) Adapted supervised change detection}} \\
    ObjFormer & 43.56 / 48.50 / 45.90 & 74.73 / 58.30 / 65.50 & 49.40 / 48.80 / 49.10 \\
    SEIFNet & 63.16 / 56.60 / 59.70 & 77.06 / 64.80 / 70.40 & 69.37 / 66.30 / 67.80 \\
    SNUNet-CD & 54.50 / 49.90 / 52.10 & 70.30 / 48.50 / 57.40 & 66.31 / 67.70 / 67.00 \\
    WDMF-Net & 51.82 / 53.40 / 52.60 & 49.17 / 51.90 / 50.50 & 49.40 / 49.40 / 49.40 \\
    ChangeFormer & 54.25 / 49.20 / 51.60 & 65.62 / 64.20 / 64.90 & 73.62 / 64.40 / 68.70 \\
    ChangeFormer-Pair & 68.84 / 37.64 / 48.67 & 55.63 / 67.23 / 60.89 & 80.28 / 36.98 / 50.64 \\
    HAM-CD & 58.07 / 53.70 / 55.80 & 49.10 / 48.70 / 48.90 & 64.56 / 69.20 / 66.80 \\
    HAM-CD-Pair & 50.39 / 65.76 / 57.06 & 45.13 / 21.70 / 29.31 & 66.28 / 55.04 / 60.14 \\
    CDMamba & 54.18 / 51.30 / 52.70 & 50.15 / 52.50 / 51.30 & 49.54 / 51.50 / 50.50 \\
    CDMamba-Pair & 59.91 / 32.04 / 41.75 & 64.44 / 49.36 / 55.90 & 64.76 / 44.81 / 52.97 \\
    \midrule
    \textbf{UniMap-Verify } & 60.28 / 66.37 / \textbf{63.18} & 81.36 / 73.21 / \textbf{77.07} & 81.02 / 75.96 / \textbf{78.41} \\
    \bottomrule
  \end{tabular*}
\par%
\par}

\end{table}

\subsection{Results}
\label{subsec:results}

\begin{figure*}[!t]
    \centering
    \includegraphics[width=0.8\textwidth]{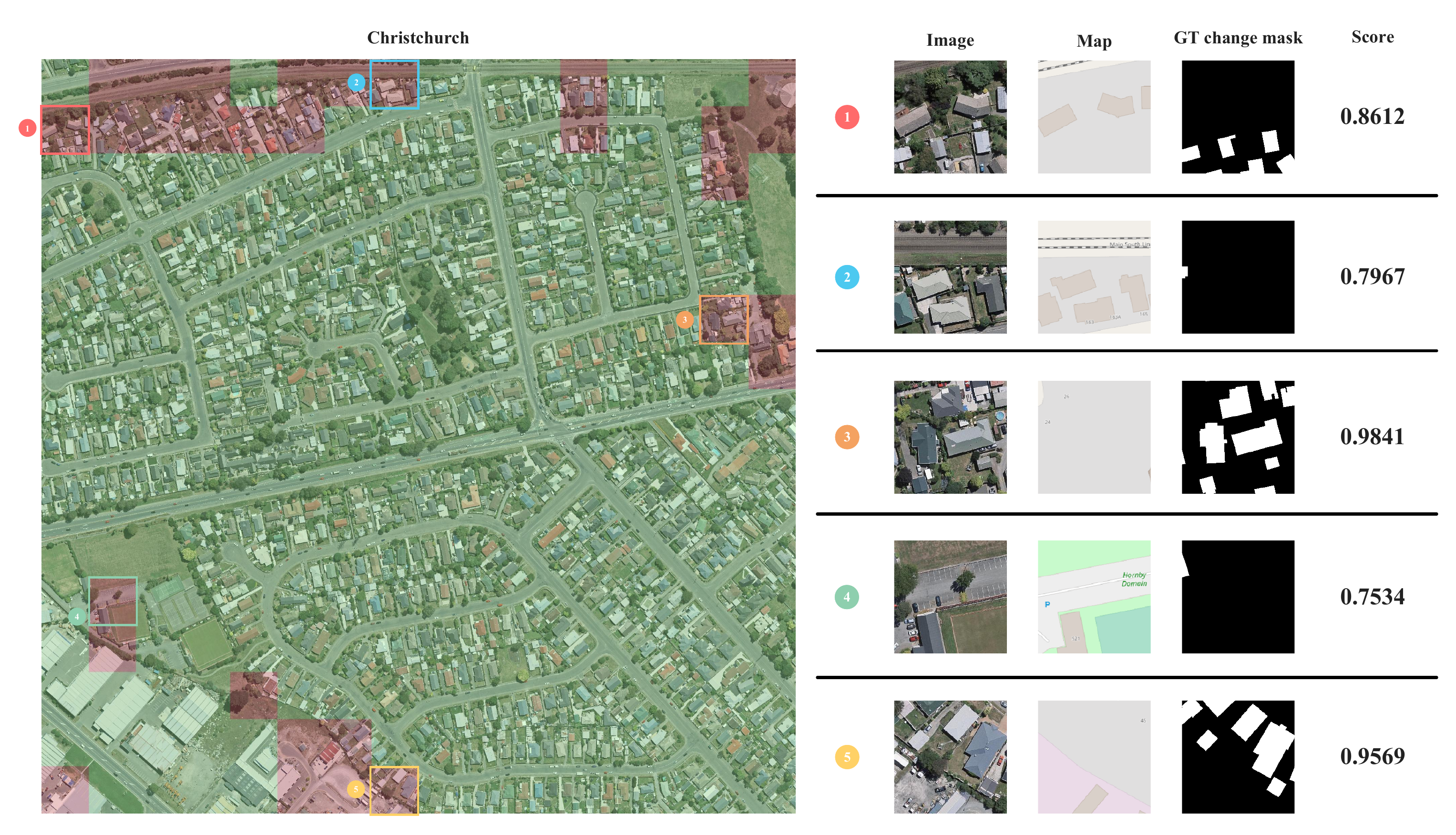}
    \caption{Large-area map verification in Christchurch. Green and red overlays denote predicted consistent and inconsistent regions, respectively. The right panel shows representative locations with their aerial image, map, ground-truth change mask, and image-level inconsistency score.}
    \label{fig:christchurch_scores}
\end{figure*}

As Figure~\ref{fig:results} illustrates, UniMap-Verify effectively identifies change regions between maps and aerial images. Compared with competing methods, it localizes true changes more accurately while suppressing spurious responses in unchanged regions, demonstrating superior localization capability and robustness. Moreover, Figures~\ref{fig:bigcity} and \ref{fig:christchurch_scores} showcase the model's capability to process large-scale urban areas in different geographic settings. Table~\ref{tab:main_results} summarizes the quantitative comparison results under the one-class, positive-pair-only protocol.

\rone{\textbf{Comparison with unsupervised and one-class methods.} As shown in Table~\ref{tab:main_results}, UniMap-Verify consistently outperforms the unsupervised heterogeneous change detection, one-class anomaly detection, and cross-modal self-supervised baselines on all three datasets. It achieves AUROCs of 64.82\%, 80.77\%, and 80.20\% and AUPRs of 67.88\%, 82.97\%, and 80.59\% on Inria, WHU, and OpenMapCD, respectively. Compared with the strongest baseline across these three categories for each dataset and metric, UniMap-Verify improves AUROC by 6.38, 15.93, and 12.93 percentage points and AUPR by 6.77, 15.43, and 16.97 percentage points, respectively. The gains are particularly pronounced on WHU and OpenMapCD, where the improvements exceed 12 percentage points in AUROC and 15 percentage points in AUPR. These consistent improvements across both metrics support the effectiveness of our task-focused consistency learning framework under positive-pair-only training.}

\rone{\textbf{Comparison with adapted supervised architectures.} Without pair-level adaptation, the strongest AUROCs are 53.49\% on Inria (HAM-CD), 59.92\% on WHU (CDMamba), and 61.04\% on OpenMapCD (ChangeFormer), all below UniMap-Verify. The pair-level variants provide an additional comparison in which supervised architectures learn from matched pairs and randomly constructed mismatches. The strongest pair-level AUROCs are 61.76\% on Inria (ChangeFormer-Pair), 65.49\% on WHU (CDMamba-Pair), and 68.60\% on OpenMapCD (HAM-CD-Pair). UniMap-Verify exceeds these values by 3.06, 15.28, and 11.60 percentage points, respectively. These gains demonstrate the effectiveness of UniMap-Verify against the evaluated pair-level adaptations under the same positive-pair-only training protocol.}

\subsection{Geographically Disjoint Evaluation on OpenMapCD-Split}
\label{subsec:geographic_evaluation}

To evaluate geographic generalization under a stricter protocol, we reorganize OpenMapCD into a geographically disjoint split, denoted as \emph{OpenMapCD-Split}. Samples from Europe and North America are used for training and validation, while samples from Asia (including Australia), South America, and Africa are reserved for region-wise testing. All methods are retrained and evaluated on this split following the same positive-pair-only protocol as the main experiments.

\rone{As shown in Table~\ref{tab:openmapcd_split_results}, UniMap-Verify achieves AUROCs of 79.08\%, 89.93\%, and 76.70\% in Asia (including Australia), South America, and Africa. It exceeds the strongest baseline AUROC in each region by 8.13, 3.35, and 13.82 percentage points, respectively. Its AUPR values are 86.21\%, 97.37\%, and 65.06\%, respectively, also exceeding the strongest baseline in each region by 7.23, 2.35, and 5.15 percentage points. These consistent region-wise results show that the advantage of UniMap-Verify persists when evaluation extends beyond the original geographic domain to unseen continental regions.}

\subsection{Threshold-Based Evaluation}
\label{subsec:threshold_evaluation}

In practical map verification, the continuous anomaly score must be converted into a binary decision at a specified operating point. An image--map pair is classified as changed when $A(x) \geq \tau$, where $\tau$ denotes the decision threshold.

\rone{For every compared method, the threshold is selected independently for each dataset exclusively on the validation set by maximizing Youden's $J$ statistic, the difference between the true positive rate (TPR) and the false positive rate (FPR). The selection rule and validation split are shared across methods.} The selected threshold is then fixed and applied to the corresponding test set without further adjustment. \rone{Table~\ref{tab:threshold_results} reports test-set Precision (Pr), Recall (Re), and F1 for the changed class.}

\rone{UniMap-Verify achieves the highest F1 on all three datasets (63.18\%, 77.07\%, and 78.41\% on Inria, WHU, and OpenMapCD, respectively). UniMap-Verify has the highest Precision on WHU and OpenMapCD, while ChangeFormer-Pair has the highest Precision on Inria. The highest Recall is obtained by UniMap-Verify on Inria and WHU, and by PatchCore on OpenMapCD. The highest F1 on all three datasets demonstrates that UniMap-Verify achieves the best precision--recall balance among the compared methods under the shared validation-based threshold-selection rule.}

\subsection{Ablation Study}
\label{subsec:ablation_study}

\begin{table*}[t]
\rone{%
\centering\footnotesize
\caption{Prototype construction (AUROC/AUPR, \%). Prototype-only uses the reference branch; Full model combines both branch scores. Best values within each column are bold.}
\label{tab:prototype_construction}
\setlength{\tabcolsep}{3pt}\renewcommand{\arraystretch}{1.08}
\begin{tabular*}{\textwidth}{@{\extracolsep{\fill}}lcccccc@{}}
\toprule
Construction & \multicolumn{2}{c}{Inria} & \multicolumn{2}{c}{WHU} & \multicolumn{2}{c}{OpenMapCD} \\
\cmidrule(lr){2-3}\cmidrule(lr){4-5}\cmidrule(lr){6-7}
 & Prototype-only & Full model & Prototype-only & Full model & Prototype-only & Full model \\
\midrule
Map & \textbf{61.83} / \textbf{62.31} & \textbf{64.82} / \textbf{67.88} & \textbf{74.08} / \textbf{74.19} & 80.77 / \textbf{82.97} & \textbf{78.53} / \textbf{71.93} & 80.20 / \textbf{80.59} \\
Image & 61.76 / 62.19 & 64.80 / 67.84 & 72.22 / 71.88 & 80.88 / 82.47 & 77.04 / 69.26 & 80.16 / 80.42 \\
Joint & 61.79 / 62.22 & 64.81 / 67.86 & 72.25 / 72.32 & \textbf{80.91} / 82.56 & 77.81 / 70.55 & \textbf{80.22} / 80.53 \\
\bottomrule\end{tabular*}
\par%
\par}
\end{table*}

\begin{table}[t]
\rone{%
\centering\footnotesize
\caption{Component ablation (AUROC/AUPR, \%). PT denotes the prototype bank. Best values are bold.}
\label{tab:ablation}
\renewcommand{\arraystretch}{1.08}\setlength{\tabcolsep}{3pt}
\begin{tabular*}{\columnwidth}{@{\extracolsep{\fill}}lccc@{}}
\toprule
Configuration & Inria & WHU & OpenMapCD \\
\midrule
\textbf{Full Model} & \textbf{64.82} / \textbf{67.88} & \textbf{80.77} / \textbf{82.97} & \textbf{80.20} / \textbf{80.59} \\
w/o MLLM & 63.66 / 67.09 & 78.90 / 82.26 & 78.62 / 79.01 \\
w/o PT & 61.28 / 64.55 & 77.92 / 79.63 & 72.62 / 73.62 \\
w/o Curriculum & 63.20 / 66.62 & 79.18 / 81.92 & 79.52 / 80.14 \\
\bottomrule\end{tabular*}
\par%
\par}
\end{table}

\rone{Table~\ref{tab:ablation} reports three ablations that separately remove MLLM-guided negative sampling, the conditional prototype bank, and the curriculum schedule to assess their contributions. The w/o MLLM configuration uses random negative sampling; w/o PT uses discriminator-only inference; w/o Curriculum removes the progressive sampling schedule. All results use the validation-set normalization in Eq.~\eqref{eq:score_normalization}, and performance drops below are measured relative to the full model in percentage points. Removing MLLM guidance reduces AUROC/AUPR by 1.16/0.79, 1.87/0.71, and 1.58/1.58 percentage points on Inria, WHU, and OpenMapCD, respectively.}

\rone{The w/o MLLM variant still exceeds the highest baseline AUROC in Table~\ref{tab:main_results} by 1.90, 13.41, and 10.02 percentage points, respectively, showing that random negatives support effective verification while semantic guidance yields further gains. Section~\ref{subsec:sampling_strategies} further analyzes the contribution of MLLM guidance by comparing different negative-sampling strategies.}

\rone{Removing the prototype bank reduces AUROC/AUPR by 3.54/3.33 on Inria, 2.85/3.34 on WHU, and 7.58/6.97 on OpenMapCD. The larger gain on OpenMapCD might be due to its greater variation in building layouts and image appearance across cities~\cite{luo2022crossspatiotemporal}. Such variation may make normal map--image correspondences harder for a single discriminator to model. PT uses each map feature to select an associated image prototype and compares the observed image feature with this reference, providing an explicit map-conditioned comparison that may be more useful in diverse scenes. WHU's single-city setting may contain more uniform correspondences that the discriminator can model more effectively. The improvement in both metrics on every dataset supports using the reference branch as a complementary source of evidence. }

\rone{Compared with the full model, w/o Curriculum shows AUROC/AUPR drops of 1.62/1.26, 1.59/1.05, and 0.68/0.45 percentage points on Inria, WHU, and OpenMapCD, respectively. This variant retains semantic guidance and the prototype bank, allowing the comparison to assess the contribution of progressively increasing negative difficulty. The consistent drops support a training schedule that first establishes basic matched--mismatched discrimination and then increases the emphasis on harder negatives. Together, the three ablations show that negative selection, inference-time reference evidence, and the training schedule each contribute to the full model's performance.}

\begin{table}[t]
\rone{%
\centering\footnotesize
\caption{Number of prototypes (AUROC/AUPR, \%). Map-based construction; scores are from the full model. Best values are bold.}
\label{tab:prototype_capacity}
\renewcommand{\arraystretch}{1.08}\setlength{\tabcolsep}{3pt}
\begin{tabular*}{\columnwidth}{@{\extracolsep{\fill}}lccc@{}}
\toprule
Configuration & Inria & WHU & OpenMapCD \\
\midrule
$K=32$ & 64.79 / 67.85 & 80.58 / 82.73 & 80.19 / 80.58 \\
$K=64$ & 64.81 / 67.81 & 80.75 / 82.87 & 80.20 / 80.59 \\
$K=128$ & \textbf{64.82} / 67.88 & \textbf{80.77} / \textbf{82.97} & 80.20 / 80.59 \\
$K=256$ & 64.78 / 67.86 & 80.74 / 82.71 & 80.21 / 80.60 \\
$K=512$ & 64.80 / \textbf{67.89} & 80.71 / 82.65 & \textbf{80.23} / \textbf{80.62} \\
\bottomrule\end{tabular*}
\par%
\par}
\end{table}

\subsection{\rone{Prototype Bank Analysis}}
\label{subsec:prototype_bank_analysis}

\subsubsection{\rone{Number of Prototypes}}
\label{subsec:prototype_analysis}

\rone{The capacity analysis varies the number of prototype slots over $K=32,64,128,256,512$ using map-based construction. Table~\ref{tab:prototype_capacity} reports the full-model scores. Here, $K$ is the number of candidate slots; only slots assigned training samples are retained for retrieval. Across the tested values, AUROC spans 0.04, 0.19, and 0.04 points on Inria, WHU, and OpenMapCD, respectively, and AUPR spans 0.08, 0.32, and 0.04 points. The limited variation indicates low sensitivity within this range. These results support $K=128$ as a common setting that maintains stable performance across all three datasets without dataset-specific adjustment of the prototype count.}

\subsubsection{\rone{Clustering Strategies}}

\rone{Following Eqs.~\eqref{eq:pt_key_assignment}--\eqref{eq:pt_value}, we cluster Map, Image, or Joint features to construct prototypes. We compare their performance under PT-only and Full inference. As shown in Table~\ref{tab:prototype_construction}, Map achieves the highest PT-only AUROC and AUPR on all three datasets, whereas the Full results differ only slightly across construction choices. Map features emphasize building footprints and their spatial arrangement, while Image and Joint features also contain visual appearance information. Grouping matched samples by map structure gives the associated image features a common structural basis, allowing their mean to summarize normal appearances for similar building patterns. PT-only relies entirely on these references, so it directly reflects differences in their usefulness for verification; in Full, the reference score is combined with the learned verification score according to Eq.~\eqref{eq:score_fusion}, and the smaller performance differences show that the final prediction is less dependent on how the references are constructed. We adopt Map for prototype construction.}

\subsection{\rone{Negative Sampling Analysis}}

\label{subsec:semantic_analysis}

\begin{figure}[!t]
\centering\includegraphics[width=\columnwidth]{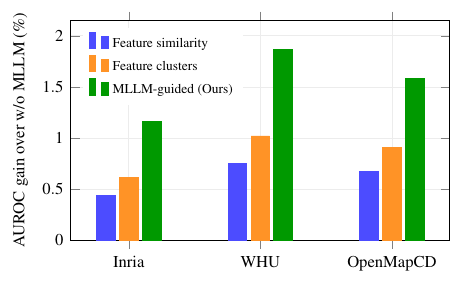}
\caption{\rone{AUROC gains (\%) over w/o MLLM for feature-similarity sampling, feature-cluster sampling, and MLLM-guided sampling (Ours). The reference uses random negative sampling without MLLM guidance.}}
\label{fig:sampling_analysis}
\end{figure}

\subsubsection{\rone{Sampling Strategies}}
\label{subsec:sampling_strategies}

\rone{To evaluate the contribution of MLLM guidance, we compare four negative-sampling strategies on the three datasets: random sampling (w/o MLLM), feature-similarity sampling, feature-cluster sampling, and MLLM-guided sampling (Ours). The w/o MLLM configuration forms negatives by random cross-sample pairing without semantic guidance. Feature-similarity sampling and feature-cluster sampling provide two alternatives based on visual features, while our MLLM-guided strategy uses the categorical semantic groups described in Section III-B. Figure~\ref{fig:sampling_analysis} reports AUROC gains over w/o MLLM, so a zero gain denotes the random-sampling reference. Both feature-based alternatives improve on this reference. MLLM guidance further improves AUROC/AUPR over feature-cluster sampling by 0.55/0.38, 0.85/0.29, and 0.67/0.71 percentage points on Inria, WHU, and OpenMapCD, respectively. The consistent gains over both feature-based alternatives support our choice of MLLM-guided sampling for improving map--image verification.}

\subsubsection{\rone{Label Noise}}
\label{subsec:label_noise}

\rone{To assess sensitivity to semantic label noise, we evaluate perturbations of the MLLM-generated labels used for negative sampling at rates of 0\%, 10\%, 20\%, 40\%, 60\%, 80\%, and 100\%. The perturbation rate is the proportion of training samples whose semantic labels are modified. We retrain the model under each perturbation setting using the resulting labels for negative sampling. Table~\ref{tab:semantic_perturbation} reports AUROC and AUPR on Inria, WHU, and OpenMapCD. The 0\% perturbation setting is the unperturbed reference; w/o MLLM is a separate reference that uses random negative sampling without semantic guidance. At 20\% perturbation, AUROC decreases by 0.20, 0.38, and 0.36 percentage points relative to the 0\% perturbation setting, respectively. At 60\% perturbation, both metrics remain above w/o MLLM on all three datasets. At 100\% perturbation, AUROC decreases by 1.46, 2.24, and 2.03 percentage points relative to the 0\% perturbation setting, and both metrics fall below w/o MLLM. These results demonstrate robustness to the tested moderate semantic-label perturbations, supporting the use of imperfect MLLM labels for negative sampling within this range.}

\begin{table}[t]

\rone{%
\centering\footnotesize
\caption{Semantic-label perturbation (AUROC/AUPR, \%). The w/o MLLM reference uses random negative sampling without semantic guidance. Best values are bold.}
\label{tab:semantic_perturbation}
\setlength{\tabcolsep}{3pt}
\begin{tabular}{lccc}
\toprule
Perturbation & Inria & WHU & OpenMapCD \\
\midrule
0\% & \textbf{64.82 / 67.88} & \textbf{80.77 / 82.97} & \textbf{80.20 / 80.59} \\
10\% & 64.76 / 67.82 & 80.62 / 82.90 & 80.05 / 80.43 \\
20\% & 64.62 / 67.68 & 80.39 / 82.80 & 79.84 / 80.22 \\
40\% & 64.30 / 67.43 & 79.97 / 82.59 & 79.44 / 79.88 \\
60\% & 63.99 / 67.20 & 79.47 / 82.39 & 79.04 / 79.45 \\
80\% & 63.65 / 67.03 & 78.93 / 82.23 & 78.55 / 78.98 \\
100\% & 63.36 / 66.87 & 78.53 / 82.02 & 78.17 / 78.69 \\
\midrule
w/o MLLM & 63.66 / 67.09 & 78.90 / 82.26 & 78.62 / 79.01 \\
\bottomrule
\end{tabular}
\par%
\par}

\end{table}

\subsubsection{\rone{Feature Distributions}}

\begin{figure*}[!t]
  \centering
  \includegraphics[width=0.98\textwidth]{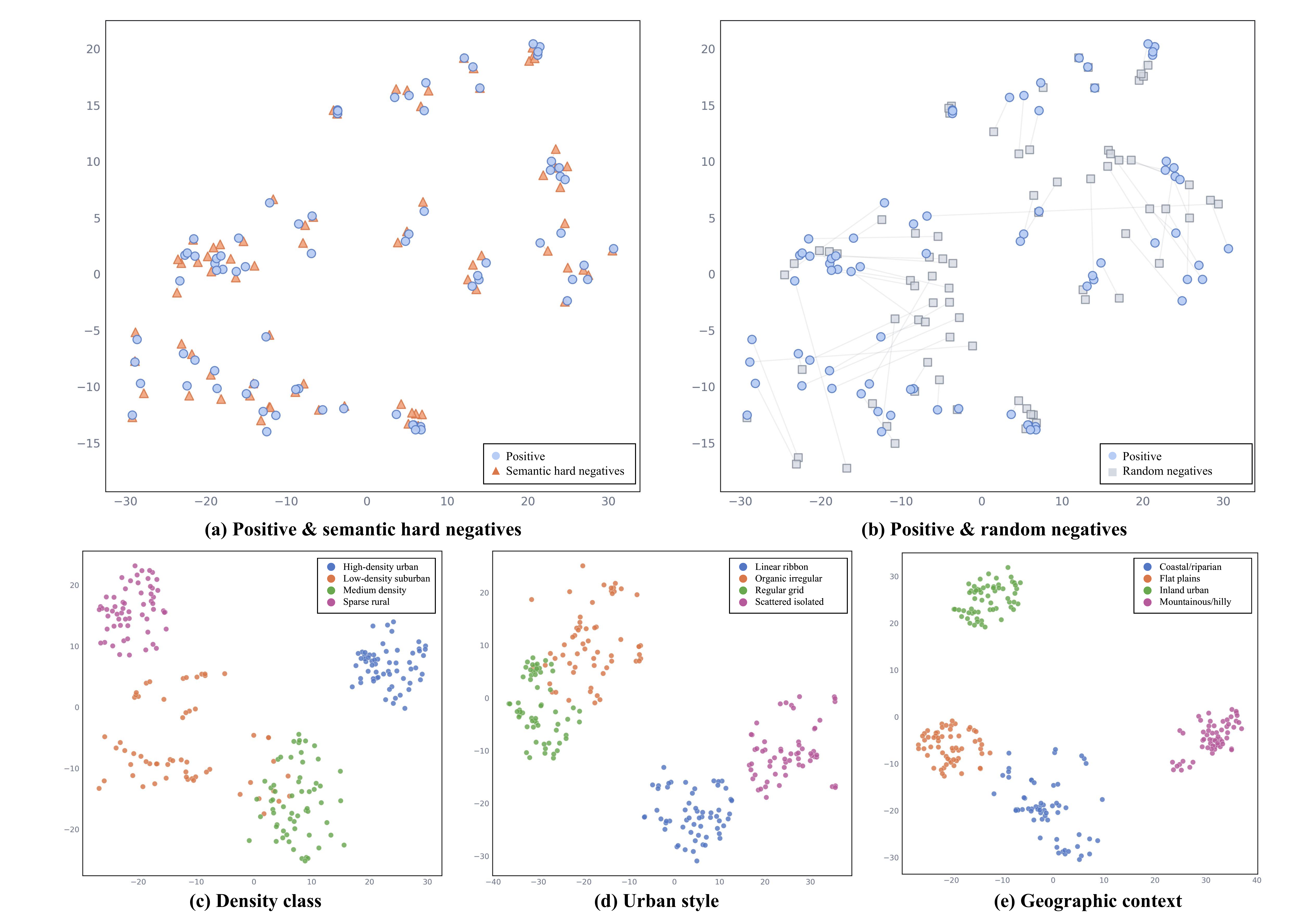}
  \caption{\textbf{Latent-space analysis of hard-negative construction on the OpenMapCD training set.} (a) t-SNE projection of positive pairs and their selected semantic hard negatives. (b) Projection of the same positive anchors and randomly sampled negatives. (c)--(e) Projections colored by the MLLM-derived density class, urban style, and geographic context, respectively. Semantic hard negatives remain locally close to their positive anchors, whereas random negatives are more dispersed; the three semantic attributes also induce structured groups in the learned representation space.}
  \label{fig:tsne_hard_negatives}
\end{figure*}

\rone{To examine whether the MLLM-defined semantic groups produce difficult negatives in the learned representation space, we extract joint embeddings from the OpenMapCD training set and visualize them using t-SNE. As shown in Figure~\ref{fig:tsne_hard_negatives}(a), semantic hard negatives exhibit substantial local overlap with their positive anchors, whereas randomly sampled negatives are more dispersed around the same positive set in Figure~\ref{fig:tsne_hard_negatives}(b). This contrast suggests that within-group sampling suppresses trivial semantic mismatches and yields negatives that are more confusable with matched pairs, which is consistent with the intended shift from coarse semantic separation toward fine-grained geometric verification. Figures~\ref{fig:tsne_hard_negatives}(c)--(e) further show that the three MLLM-derived attributes---density class, urban style, and geographic context---form structured groups in the learned space rather than acting as arbitrary identifiers.}

\subsubsection{\rone{Change Size}}

\label{subsec:change_size}

\begin{figure*}[!t]
\centering
\setlength{\abovecaptionskip}{3pt}
\includegraphics[width=\textwidth,trim=0 5bp 0 0,clip]{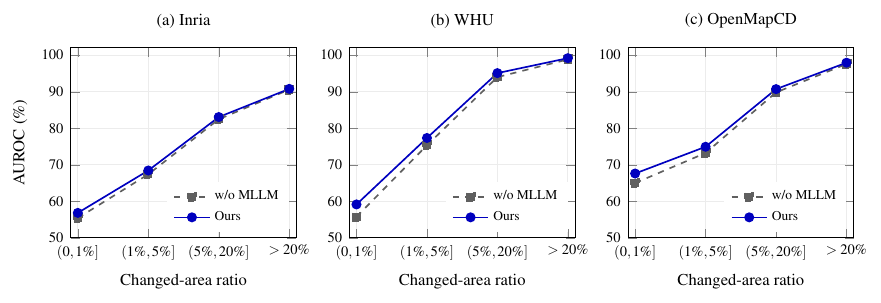}\par
\caption{\rone{Change-size sensitivity of UniMap-Verify with MLLM sampling (Ours) and random negative sampling without MLLM guidance (w/o MLLM) on (a) Inria, (b) WHU, and (c) OpenMapCD. AUROC is reported for ground-truth changed-area groups; area intervals are categorical bins.}}
\label{fig:score_vs_change_size}
\end{figure*}

\rone{The change-size analysis examines whether semantic guidance also benefits verification of small real changes. We compare the full model with w/o MLLM across ground-truth changed-area groups on each dataset. The changed-area ratio is the fraction of an image covered by its ground-truth change mask; the groups are $(0,1\%]$, $(1\%,5\%]$, $(5\%,20\%]$, and $>20\%$. Figure~\ref{fig:score_vs_change_size} reports AUROC for these groups. MLLM-guided sampling achieves higher AUROC in every group. For changes covering at most 1\% of an image, AUROC increases from 55.43\% to 56.85\% on Inria, from 55.78\% to 59.16\% on WHU, and from 65.01\% to 67.62\% on OpenMapCD, giving gains of 1.42, 3.38, and 2.61 percentage points. Although the smallest changes remain more difficult, the consistent gains demonstrate that semantic-guided training improves verification of real local inconsistencies across the tested change sizes.}

\begin{table}[t]
\centering
    \caption{\textbf{Statistical analysis of generated semantic descriptors on the Inria dataset.} While all models adhere strictly to the output schema, Qwen3-VL-8B exhibits significantly higher lexical diversity and description length.}
\label{tab:text_stats}
    \resizebox{\columnwidth}{!}{
    \begin{tabular}{l|c|c|c}
        \toprule
        \textbf{Model} & \textbf{Valid JSON (\%)} & \textbf{Avg. Length (words)} & \textbf{Vocab Size} \\
        \midrule
        InternVL3-8B & 100.00\% & 22.45 & 509 \\
        Qwen2.5-VL-3B   & 100.00\% & 19.97 & 178 \\
        Qwen2.5-VL-7B   & 100.00\% & 18.84 & 436 \\
        \textbf{Qwen3-VL-8B} & \textbf{100.00\%} & \textbf{24.16} & \textbf{1,206} \\
        \bottomrule
    \end{tabular}
    }
\par
\end{table}

\subsubsection{Semantic Descriptors}

\label{subsec:llm_perf}\label{subsec:text_stats}\label{subsec:why_qwen3vl}

\begin{figure*}[!t]
    \centering
    \includegraphics[width=1\linewidth]{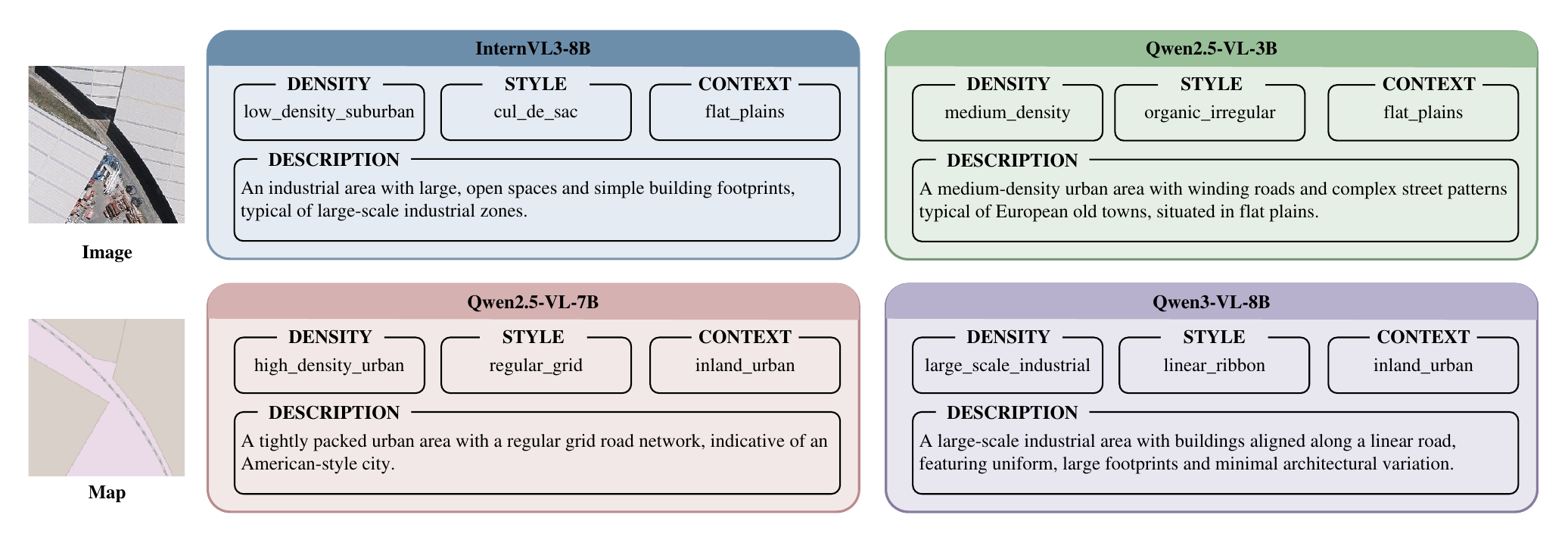}
    \caption{Comparison of MLLM-generated semantic descriptions. We compare InternVL3-8B, Qwen2.5-VL-3B, Qwen2.5-VL-7B, and Qwen3-VL-8B on a sample pair.}
    \label{fig:mllm_comparison}
\end{figure*}

Our Semantic-Guided Hard Negative Mining relies on MLLM-generated semantic descriptors to cluster samples and construct semantically similar but geometrically mismatched negatives. We therefore compare four candidate MLLMs qualitatively: InternVL3-8B, Qwen2.5-VL-3B, Qwen2.5-VL-7B, and Qwen3-VL-8B. As illustrated in Figure~\ref{fig:mllm_comparison}, Qwen3-VL-8B tends to produce more specific and spatially grounded descriptors, whereas the smaller models more often collapse to generic urban labels. 

Beyond qualitative inspection, we directly analyzed the quality of the generated JSONL descriptors on the Inria dataset (24,380 samples). Table~\ref{tab:text_stats} summarizes three metrics: (1) \textbf{Schema Compliance} (rate of valid JSON parsing), (2) \textbf{Descriptive Detail} (average word count per description), and (3) \textbf{Lexical Diversity} (unique vocabulary size).

As summarized in Table~\ref{tab:text_stats}, we observed the following:
\begin{itemize}
    \item \textbf{High Schema Adherence:} All four models achieved a 100\% valid JSON rate, indicating that our prompt template is robust and compatible across different MLLM architectures.
    \item \textbf{Superior Lexical Diversity:} The most striking difference lies in vocabulary size. Qwen3-VL-8B utilizes a vocabulary of 1,206 unique words, which is nearly 2.4$\times$ that of InternVL3-8B (509) and 6.7$\times$ that of Qwen2.5-VL-3B (178). This suggests that smaller or less capable models tend to reuse a limited set of generic terms (e.g., repeating "dense residential" for all city blocks), whereas Qwen3-VL-8B uses a much richer vocabulary to describe subtle variations in urban texture, vegetation, and style.
    \item \textbf{Richness of Detail:} Qwen3-VL-8B also generates the longest descriptions (24.16 words/sample), providing more detailed context than Qwen2.5-VL-7B (18.84 words), which further supports the effectiveness of our semantic-guided hard negative mining.
\end{itemize}

\begin{table}[!t]
\rone{%
\centering\footnotesize
\caption{Comparison of normalization strategies (AUROC/AUPR, \%). Best values are bold.}
\label{tab:normalization}
\renewcommand{\arraystretch}{1.08}\setlength{\tabcolsep}{1.5pt}
\begin{tabular*}{\columnwidth}{@{\extracolsep{\fill}}lccc@{}}
\toprule
Configuration & Inria & WHU & OpenMapCD \\
\midrule
Per-pair normalization & 64.69/66.72 & 77.79/80.39 & 77.53/77.29 \\
No normalization & 64.33/67.51 & 80.70/82.04 & 80.05/80.40 \\
Validation-set normalization & \textbf{64.82}/\textbf{67.88} & \textbf{80.77}/\textbf{82.97} & \textbf{80.20}/\textbf{80.59} \\
\bottomrule\end{tabular*}
\par%
\par}
\end{table}

Based on the qualitative comparison in Figure~\ref{fig:mllm_comparison} and the text-level statistics in Table~\ref{tab:text_stats}, we selected \textbf{Qwen3-VL-8B} as the default generator for the following reasons:
\begin{enumerate}
    \item \textbf{Richer Semantic Specificity:} It yields the largest vocabulary and the longest descriptions, indicating better coverage of subtle urban attributes that are useful for semantic grouping.
    \item \textbf{No Robustness Trade-off:} It maintains the same 100\% valid-JSON rate as the other candidates, so the added descriptive richness does not come at the cost of schema reliability.
\end{enumerate}

\begin{table}[!t]
\rone{%
    \centering
    \caption{Sensitivity analysis of $\alpha$ on the WHU dataset.}
    \label{tab:alpha_sensitivity}
    \resizebox{0.48\textwidth}{!}{
    \begin{tabular}{l|ccccc|c|ccc}
        \toprule
        $\alpha$ & 0.0 & 0.1 & 0.3 & 0.5 & 0.6 & \textbf{0.7} & 0.8 & 0.9 & 1.0 \\
        \midrule
        Branch & \multicolumn{5}{c|}{Prototype-Dominant} & \textbf{Fused} & \multicolumn{3}{c}{Discriminator-Dominant} \\
        \midrule
        AUROC & 74.08 & 76.03 & 79.00 & 80.19 & 80.73 & \textbf{80.77} & 79.74 & 78.66 & 77.92 \\
        \bottomrule
    \end{tabular}
    }
\par%
\par}
\end{table}

\subsection{\rone{Normalization}}
\label{subsec:score_calibration}

\rone{We compare different normalization strategies applied before combining the discriminator and prototype scores in Table~\ref{tab:normalization}. These strategies differ in how the two branch scores are scaled before weighted fusion. Per-pair normalization scales each branch using statistics from the current input pair, then combines the normalized scores. No normalization directly combines the raw branch scores. Validation-set normalization scales each branch using fixed statistics computed on the validation set, as in Eq.~\eqref{eq:score_normalization}, before combining the scores according to Eq.~\eqref{eq:score_fusion}. Per-pair normalization removes absolute score offsets, reducing comparability across input pairs. Validation-set normalization preserves a common scale for combining the two branch scores. It achieves the highest AUROC and AUPR on all three datasets (Table~\ref{tab:normalization}) and is therefore adopted for the component and prototype analyses.}

\subsection{\rone{Fusion Weight}}
\label{subsec:fusion_weight}

\rone{The fusion weight $\alpha$ in Eq.~\eqref{eq:score_fusion} controls the relative contributions of the two scores. We vary this weight on WHU to evaluate its effect on inference. The endpoint $\alpha=0$ uses only the prototype score, and $\alpha=1$ uses only the discriminator score; intermediate values combine them before spatial maximum pooling. Table~\ref{tab:alpha_sensitivity} shows the highest measured AUROC of 80.77\% at $\alpha=0.7$, compared with 74.08\% and 77.92\% at the two endpoints. The combined score therefore gains 6.69 and 2.85 percentage points over prototype-only and discriminator-only inference, respectively. The intermediate settings $\alpha=0.2$ and $0.4$ obtain 77.56\% and 79.94\%. These results demonstrate the complementary value of the two branch scores and support weighted score fusion, with $\alpha=0.7$ selected for the evaluated WHU setting.}

\section{Conclusion}
  Keeping maps consistent with the evolving physical world is crucial for modern navigation, yet remains costly to maintain. We introduced   UniMap-Verify for unsupervised / one-class map-image consistency verification under a strict positive-pair-only training protocol. The central contribution is MLLM-guided semantic hard negative mining, together with a curriculum schedule that progressively increases training difficulty; the prototype bank provides a complementary normal-reference cue. We also presented MapCheck-Bench, a large-scale benchmark with rich semantic annotations to facilitate rigorous evaluation, including approximately 47k map-image-text triplets across 42 unique cities/regions. Extensive experiments on the proposed benchmark demonstrate that UniMap-Verify substantially outperforms existing unsupervised baselines in this one-class setting. While this work focuses on building-footprint verification, the semantic hard-negative framework is generic and can be extended to other map elements such as roads, water systems, or land use. 
\rone{In routine map maintenance, UniMap-Verify could compare each map tile with its corresponding aerial image and rank the tiles by inconsistency score, helping human reviewers decide which areas to inspect first. In post-disaster assessment, it could compare pre-disaster maps with post-disaster aerial imagery to identify areas where buildings may have been destroyed.}

\bibliography{example_paper}

\begin{thebibliography}{10}
\providecommand{\url}[1]{#1}
\csname url@samestyle\endcsname
\providecommand{\newblock}{\relax}
\providecommand{\bibinfo}[2]{#2}
\providecommand{\BIBentrySTDinterwordspacing}{\spaceskip=0pt\relax}
\providecommand{\BIBentryALTinterwordstretchfactor}{4}
\providecommand{\BIBentryALTinterwordspacing}{\spaceskip=\fontdimen2\font plus
\BIBentryALTinterwordstretchfactor\fontdimen3\font minus
  \fontdimen4\font\relax}
\providecommand{\BIBforeignlanguage}[2]{{%
\expandafter\ifx\csname l@#1\endcsname\relax
\typeout{** WARNING: IEEEtran.bst: No hyphenation pattern has been}%
\typeout{** loaded for the language `#1'. Using the pattern for}%
\typeout{** the default language instead.}%
\else
\language=\csname l@#1\endcsname
\fi
#2}}
\providecommand{\BIBdecl}{\relax}
\BIBdecl

\bibitem{fesenmyer2025addressing}
K.~A. Fesenmyer, E.~E. Poor, D.~E. Terasaki~Hart, J.~W. Veldman, F.~Fleischman,
  P.~Choksi, S.~Archibald, M.~Armani, M.~E. Fagan, E.~C. Fricke \emph{et~al.},
  ``Addressing critiques refines global estimates of reforestation potential
  for climate change mitigation,'' \emph{Nature communications}, vol.~16,
  no.~1, p. 4572, 2025.

\bibitem{ding2020practical}
L.~Ding, J.~Zhou, L.~Meng, and Z.~Long, ``A practical cross-view image matching
  method between {UAV} and satellite for {UAV}-based geo-localization,''
  \emph{Remote Sensing}, vol.~13, no.~1, p.~47, 2020.

\bibitem{zhan2017change}
Y.~Zhan, K.~Fu, M.~Yan, X.~Sun, H.~Wang, and X.~Qiu, ``Change detection based
  on deep siamese convolutional network for optical aerial images,'' \emph{IEEE
  Geoscience and Remote Sensing Letters}, vol.~14, no.~10, pp. 1845--1849,
  2017.

\bibitem{Su2020BUILDINGCD}
\BIBentryALTinterwordspacing
S.-C. Su, T.~Nawata, and T.~Fuse, ``Building change detection from bitemporal
  aerial images using deep learning,'' \emph{ISPRS Annals of the
  Photogrammetry, Remote Sensing and Spatial Information Sciences}, pp.
  565--571, 2020. [Online]. Available:
  \url{https://api.semanticscholar.org/CorpusID:222076750}
\BIBentrySTDinterwordspacing

\bibitem{deng2024cross}
K.~Deng, X.~Hu, Z.~Zhang, B.~Su, C.~Feng, Y.~Zhan, X.~Wang, and Y.~Duan,
  ``Cross-modal change detection using historical land use maps and current
  remote sensing images,'' \emph{ISPRS Journal of Photogrammetry and Remote
  Sensing}, vol. 218, pp. 114--132, 2024.

\bibitem{zhong2022regionclip}
Y.~Zhong, J.~Yang, P.~Zhang, C.~Li, N.~Codella, L.~H. Li, L.~Zhou, X.~Dai,
  L.~Yuan, Y.~Li \emph{et~al.}, ``Regionclip: Region-based language-image
  pretraining,'' in \emph{Proceedings of the IEEE/CVF Conference on Computer
  Vision and Pattern Recognition}, 2022, pp. 16\,793--16\,803.

\bibitem{sun2021iterative}
Y.~Sun, L.~Lei, D.~Guan, and G.~Kuang, ``Iterative robust graph for
  unsupervised change detection of heterogeneous remote sensing images,''
  \emph{IEEE Transactions on Image Processing}, vol.~30, pp. 6277--6291, 2021.

\bibitem{dong2024contrastive}
H.~Dong, W.~Ma, L.~Jiao, F.~Liu, X.~Liu, and H.~Zhu, ``Contrastive learning
  with context-augmented transformer for change detection in {SAR} images,''
  \emph{IEEE Journal of Selected Topics in Applied Earth Observations and
  Remote Sensing}, 2024.

\bibitem{chen2024objformer}
H.~Chen, C.~Lan, J.~Song, C.~Broni-Bediako, J.~Xia, and N.~Yokoya, ``Objformer:
  Learning land-cover changes from paired {OSM} data and optical
  high-resolution imagery via object-guided transformer,'' \emph{IEEE
  Transactions on Geoscience and Remote Sensing}, vol.~62, pp. 1--22, 2024.

\bibitem{bastani2021beyond}
F.~Bastani and S.~Madden, ``Beyond road extraction: A dataset for map update
  using aerial images,'' in \emph{Proceedings of the IEEE/CVF International
  Conference on Computer Vision}, 2021, pp. 11\,905--11\,914.

\bibitem{JSTARS2022HSIChangeDetection}
X.~Ou, L.~Liu, S.~Tan, G.~Zhang, W.~Li, and B.~Tu, ``A hyperspectral image
  change detection framework with self-supervised contrastive learning
  pretrained model,'' \emph{IEEE Journal of Selected Topics in Applied Earth
  Observations and Remote Sensing}, vol.~15, pp. 7724--7740, 2022.

\bibitem{TGRS2023CSSLChangeDetection}
M.~Yang, L.~Jiao, F.~Liu, B.~Hou, S.~Yang, Y.~Zhang, and J.~Wang, ``Multicue
  contrastive self-supervised learning for change detection in remote
  sensing,'' \emph{IEEE Transactions on Geoscience and Remote Sensing},
  vol.~61, pp. 1--14, 2023.

\bibitem{CVPR2023ChangeAwareSampling}
U.~Mall, B.~Hariharan, and K.~Bala, ``Change-aware sampling and contrastive
  learning for satellite images,'' in \emph{2023 IEEE/CVF Conference on
  Computer Vision and Pattern Recognition (CVPR)}, 2023, pp. 5261--5270.

\bibitem{JSTARS2025PCLCD}
W.~Yan, C.~Zhu, M.~Wang, D.~Yu, Z.~Zou, and T.~Xia, ``Pclcd: Patch-based
  contrastive learning for remote sensing image change detection,'' \emph{IEEE
  Journal of Selected Topics in Applied Earth Observations and Remote Sensing},
  vol.~18, pp. 26\,759--26\,773, 2025.

\bibitem{pisl2021detecting}
J.~Pisl, H.~Li, S.~Lautenbach, B.~Herfort, and A.~Zipf, ``Detecting
  {OpenStreetMap} missing buildings by transferring pre-trained deep neural
  networks,'' \emph{AGILE: GIScience Series}, vol.~2, p.~39, 2021.

\bibitem{vargas2019correcting}
J.~E. Vargas-Mu{\~n}oz, S.~Lobry, A.~X. Falc{\~a}o, and D.~Tuia, ``Correcting
  rural building annotations in {OpenStreetMap} using convolutional neural
  networks,'' \emph{ISPRS journal of photogrammetry and remote sensing}, vol.
  147, pp. 283--293, 2019.

\bibitem{cherif2024novel}
M.~A. Cherif, S.~Tripodi, Y.~Tarabalka, I.~Manighetti, and L.~Laurore, ``Novel
  approaches for aligning geospatial vector maps,'' \emph{ISPRS International
  Archives of the Photogrammetry, Remote Sensing and Spatial Information
  Sciences}, vol.~48, pp. 55--64, 2024.

\bibitem{kirillov2023segment}
A.~Kirillov, E.~Mintun, N.~Ravi, H.~Mao, C.~Rolland, L.~Gustafson, T.~Xiao,
  S.~Whitehead, A.~C. Berg, W.-Y. Lo \emph{et~al.}, ``Segment anything,'' in
  \emph{Proceedings of the IEEE/CVF international conference on computer
  vision}, 2023, pp. 4015--4026.

\bibitem{mo2024survey}
S.~Mo, Y.~Shi, Q.~Yuan, and M.~Li, ``A survey of deep learning road extraction
  algorithms using high-resolution remote sensing images,'' \emph{Sensors},
  vol.~24, no.~5, p. 1708, 2024.

\bibitem{sahu2019vector}
M.~Sahu and A.~Ohri, ``Vector map generation from aerial imagery using deep
  learning,'' \emph{ISPRS Annals of the Photogrammetry, Remote Sensing and
  Spatial Information Sciences}, vol.~4, pp. 157--162, 2019.

\bibitem{chen2022unsupervised}
H.~Chen, N.~Yokoya, C.~Wu, and B.~Du, ``Unsupervised multimodal change
  detection based on structural relationship graph representation learning,''
  \emph{IEEE transactions on geoscience and remote sensing}, vol.~60, pp.
  1--18, 2022.

\bibitem{luppino2021deep}
L.~T. Luppino, M.~Kampffmeyer, F.~M. Bianchi, G.~Moser, S.~B. Serpico,
  R.~Jenssen, and S.~N. Anfinsen, ``Deep image translation with an
  affinity-based change prior for unsupervised multimodal change detection,''
  \emph{IEEE Transactions on Geoscience and Remote Sensing}, vol.~60, pp.
  1--22, 2021.

\bibitem{wang2024cd}
J.-J. Wang, N.~Dobigeon, M.~Chabert, D.-C. Wang, T.-Z. Huang, and J.~Huang,
  ``Cd-gan: A robust fusion-based generative adversarial network for
  unsupervised remote sensing change detection with heterogeneous sensors,''
  \emph{Information Fusion}, vol. 107, p. 102313, 2024.

\bibitem{fuller2023croma}
A.~Fuller, K.~Millard, and J.~Green, ``Croma: Remote sensing representations
  with contrastive radar-optical masked autoencoders,'' \emph{Advances in
  Neural Information Processing Systems}, vol.~36, pp. 5506--5538, 2023.

\bibitem{ding2025s2c}
L.~Ding, X.~Zuo, D.~Hong, H.~Guo, J.~Lu, Z.~Gong, and L.~Bruzzone, ``S2c:
  Learning noise-resistant differences for unsupervised change detection in
  multimodal remote sensing images,'' \emph{arXiv preprint arXiv:2502.12604},
  2025.

\bibitem{maggiori2017dataset}
E.~Maggiori, Y.~Tarabalka, G.~Charpiat, and P.~Alliez, ``Can semantic labeling
  methods generalize to any city? the inria aerial image labeling benchmark,''
  in \emph{IEEE International Geoscience and Remote Sensing Symposium
  (IGARSS)}.\hskip 1em plus 0.5em minus 0.4em\relax IEEE, 2017.

\bibitem{WHUBuilding}
S.~Ji, S.~Wei, and M.~Lu, ``Fully convolutional networks for multisource
  building extraction from an open aerial and satellite imagery data set,''
  \emph{IEEE Transactions on Geoscience and Remote Sensing}, vol.~57, no.~1,
  pp. 574--586, 2019.

\bibitem{xia2023openearthmap}
J.~Xia, N.~Yokoya, B.~Adriano, and C.~Broni-Bediako, ``Openearthmap: A
  benchmark dataset for global high-resolution land cover mapping,'' in
  \emph{Proceedings of the IEEE/CVF Winter Conference on Applications of
  Computer Vision}, 2023, pp. 6254--6264.

\bibitem{OpenStreetMap}
{OpenStreetMap contributors}, ``{Planet dump retrieved from
  https://planet.osm.org },'' \url{ https://www.openstreetmap.org }, 2017.

\bibitem{zagoruyko2017wideresidualnetworks}
\BIBentryALTinterwordspacing
S.~Zagoruyko and N.~Komodakis, ``Wide residual networks,'' 2017. [Online].
  Available: \url{https://arxiv.org/abs/1605.07146}
\BIBentrySTDinterwordspacing

\bibitem{Qwen3-VL}
S.~Bai, Y.~Cai, R.~Chen, K.~Chen, X.~Chen, Z.~Cheng, L.~Deng, W.~Ding, C.~Gao,
  C.~Ge, W.~Ge, Z.~Guo, Q.~Huang, J.~Huang, F.~Huang, B.~Hui, S.~Jiang, Z.~Li,
  M.~Li, M.~Li, K.~Li, Z.~Lin, J.~Lin, X.~Liu, J.~Liu, C.~Liu, Y.~Liu, D.~Liu,
  S.~Liu, D.~Lu, R.~Luo, C.~Lv, R.~Men, L.~Meng, X.~Ren, X.~Ren, S.~Song,
  Y.~Sun, J.~Tang, J.~Tu, J.~Wan, P.~Wang, P.~Wang, Q.~Wang, Y.~Wang, T.~Xie,
  Y.~Xu, H.~Xu, J.~Xu, Z.~Yang, M.~Yang, J.~Yang, A.~Yang, B.~Yu, F.~Zhang,
  H.~Zhang, X.~Zhang, B.~Zheng, H.~Zhong, J.~Zhou, F.~Zhou, J.~Zhou, Y.~Zhu,
  and K.~Zhu, ``Qwen3-vl technical report,'' \emph{arXiv preprint
  arXiv:2511.21631}, 2025.

\bibitem{wu2021commonality}
Y.~Wu, J.~Li, Y.~Yuan, A.~K. Qin, Q.-G. Miao, and M.-G. Gong, ``Commonality
  autoencoder: Learning common features for change detection from heterogeneous
  images,'' \emph{IEEE transactions on neural networks and learning systems},
  vol.~33, no.~9, pp. 4257--4270, 2021.

\bibitem{CDRL2025}
T.~Zhan, J.~Lan, Y.~Zhu, Q.~Dang, and M.~Gong, ``Cross-domain difference
  representation learning for unsupervised heterogeneous change detection,''
  \emph{IEEE Transactions on Geoscience and Remote Sensing}, vol.~63, pp.
  1--17, 2025.

\bibitem{TIP2025CFRL}
T.~Liu, M.~Zhang, M.~Gong, Q.~Zhang, F.~Jiang, H.~Zheng, and D.~Lu,
  ``Commonality feature representation learning for unsupervised multimodal
  change detection,'' \emph{IEEE Transactions on Image Processing}, vol.~34,
  pp. 1219--1233, 2025.

\bibitem{hu2023prbcd}
L.~Hu, Q.~Liu, J.~Liu, and L.~Xiao, ``Prbcd-net: Predict-refining-involved
  bidirectional contrastive difference network for unsupervised change
  detection,'' \emph{IEEE Transactions on Geoscience and Remote Sensing},
  vol.~61, pp. 1--17, 2023.

\bibitem{ruff2018deep}
\BIBentryALTinterwordspacing
L.~Ruff, R.~Vandermeulen, N.~G{\"o}rnitz, L.~Deecke, S.~A. Siddiqui, A.~Binder,
  E.~M{\"u}ller, and M.~Kloft, ``Deep one-class classification,'' in
  \emph{Proceedings of the International Conference on Machine Learning}, 2018,
  pp. 4393--4402. [Online]. Available:
  \url{https://proceedings.mlr.press/v80/ruff18a.html}
\BIBentrySTDinterwordspacing

\bibitem{roth2022total}
\BIBentryALTinterwordspacing
K.~Roth, L.~Pemula, J.~Zepeda, B.~Sch{\"o}lkopf, T.~Brox, and P.~Gehler,
  ``Towards total recall in industrial anomaly detection,'' in
  \emph{Proceedings of the IEEE/CVF Conference on Computer Vision and Pattern
  Recognition}, 2022, pp. 14\,318--14\,328. [Online]. Available:
  \url{https://openaccess.thecvf.com/content/CVPR2022/html/Roth_Towards_Total_Recall_in_Industrial_Anomaly_Detection_CVPR_2022_paper.html}
\BIBentrySTDinterwordspacing

\bibitem{tian2020contrastive}
\BIBentryALTinterwordspacing
Y.~Tian, D.~Krishnan, and P.~Isola, ``Contrastive multiview coding,'' in
  \emph{Proceedings of the European Conference on Computer Vision}, 2020.
  [Online]. Available:
  \url{https://www.ecva.net/papers/eccv_2020/papers_ECCV/html/1453_ECCV_2020_paper.php}
\BIBentrySTDinterwordspacing

\bibitem{morgado2021audio}
\BIBentryALTinterwordspacing
P.~Morgado, N.~Vasconcelos, and I.~Misra, ``Audio-visual instance
  discrimination with cross-modal agreement,'' in \emph{Proceedings of the
  IEEE/CVF Conference on Computer Vision and Pattern Recognition}, 2021, pp.
  12\,475--12\,486. [Online]. Available:
  \url{https://openaccess.thecvf.com/content/CVPR2021/html/Morgado_Audio-Visual_Instance_Discrimination_with_Cross-Modal_Agreement_CVPR_2021_paper.html}
\BIBentrySTDinterwordspacing

\bibitem{bandara2022transformer}
W.~G.~C. Bandara and V.~M. Patel, ``A transformer-based siamese network for
  change detection,'' in \emph{IGARSS 2022-2022 IEEE International Geoscience
  and Remote Sensing Symposium}.\hskip 1em plus 0.5em minus 0.4em\relax IEEE,
  2022, pp. 207--210.

\bibitem{huang2024spatiotemporal}
Y.~Huang, X.~Li, Z.~Du, and H.~Shen, ``Spatiotemporal enhancement and
  interlevel fusion network for remote sensing images change detection,''
  \emph{IEEE Transactions on Geoscience and Remote Sensing}, vol.~62, pp.
  1--14, 2024.

\bibitem{fang2021snunet}
S.~Fang, K.~Li, J.~Shao, and Z.~Li, ``Snunet-cd: A densely connected siamese
  network for change detection of vhr images,'' \emph{IEEE Geoscience and
  Remote Sensing Letters}, vol.~19, pp. 1--5, 2021.

\bibitem{liu2026lightweight}
Y.~Liu, S.~Li, and L.~Bruzzone, ``A lightweight wavelet-aligned difference and
  mask-guided fusion network for change detection,'' \emph{IEEE Transactions on
  Geoscience and Remote Sensing}, vol.~64, pp. 1--16, 2026.

\bibitem{li2026ham}
G.~Li, P.~Han, W.~Wang, T.~Mu, Z.~Xiao, and X.~Li, ``Ham-cd: Hybrid attention
  mamba for remote sensing change detection,'' \emph{IEEE Transactions on
  Geoscience and Remote Sensing}, 2026.

\bibitem{zhang2025cdmamba}
H.~Zhang, K.~Chen, C.~Liu, H.~Chen, Z.~Zou, and Z.~Shi, ``Cdmamba:
  Incorporating local clues into mamba for remote sensing image binary change
  detection,'' \emph{IEEE Transactions on Geoscience and Remote Sensing}, 2025.

\bibitem{luo2022crossspatiotemporal}
M.~Luo and S.~Ji, ``Cross-spatiotemporal land-cover classification from {VHR}
  remote sensing images with deep learning based domain adaptation,''
  \emph{ISPRS Journal of Photogrammetry and Remote Sensing}, vol. 191, pp.
  105--128, 2022.

\end{thebibliography}
\bibliographystyle{IEEEtran}

\end{document}